\documentclass[11pt]{article}

\usepackage{acl}

\usepackage{times}
\usepackage{latexsym}
\usepackage[T1]{fontenc}
\usepackage[utf8]{inputenc}
\usepackage{microtype}
\usepackage{inconsolata}
\usepackage{graphicx}
\usepackage{booktabs}
\usepackage{array}
\usepackage{tabularx}
\usepackage{makecell}
\usepackage{multirow}
\usepackage[normalem]{ulem}

\usepackage{amssymb}

\usepackage{placeins}
\usepackage{amsmath}
\usepackage[table]{xcolor}
\usepackage{listings}
\usepackage[most]{tcolorbox}
\usepackage{hyperref}

\usepackage{xspace}
\usepackage{svg}
\usepackage{tikz}
\usepackage[table]{xcolor}   
\usepackage{colortbl}        
\usepackage{fontawesome5}

\definecolor{vancolor}{RGB}{225,110,95}      
\definecolor{divcolor}{RGB}{55,145,165}      
\definecolor{humanbg}{RGB}{222,238,224}
\definecolor{humanfg}{RGB}{40,105,60}
\definecolor{nohumanbg}{RGB}{250,222,222}
\definecolor{nohumanfg}{RGB}{170,40,40}
\definecolor{secbg}{RGB}{243,245,251}
\definecolor{secfg}{RGB}{45,60,100}

\newcommand{\segbar}[3]{
  \tikz[baseline=-0.4ex]{%
    \foreach \i in {1,...,#2}{%
      \pgfmathtruncatemacro{\flag}{\i<=#1 ? 1 : 0}%
      \ifnum\flag=1
        \fill[#3] (0.66em*\i,0) circle (0.22em);
      \else
        \draw[gray!70,line width=0.8pt,fill=white] (0.66em*\i,0) circle (0.22em);
      \fi
    }%
  }%
}
\newcommand{\pill}[3]{\tikz[baseline=-0.55ex]\node[rounded corners=2.5pt, fill=#1, text=#2,
  font=\scriptsize\sffamily\bfseries, inner xsep=4pt, inner ysep=1.4pt] {#3};}
\newcommand{\humantag}[2]{%
  \,\ifnum#1=0
    \pill{nohumanbg}{nohumanfg}{LLM-only \,#1/#2}%
  \else
    \pill{humanbg}{humanfg}{human \,#1/#2}%
  \fi
}

\newcolumntype{L}[1]{>{\raggedright\arraybackslash}p{#1}}
\newcolumntype{Y}{>{\raggedright\arraybackslash}X}

\newif\ifcommentsoff
\commentsofffalse 

\lstdefinestyle{promptbox}{
  basicstyle=\ttfamily\scriptsize,
  columns=fullflexible,
  keepspaces=true,
  breaklines=true,
  breakatwhitespace=false,
  showstringspaces=false,
  frame=none,
  backgroundcolor=\color{white},
  xleftmargin=0pt,
  xrightmargin=0pt,
  aboveskip=0pt,
  belowskip=0pt,
  literate=
    {→}{{$\rightarrow$}}1
    {—}{{---}}1
    {–}{{--}}1
    {‑}{{-}}1
    {“}{{``}}1
    {”}{{''}}1
    {‘}{{`}}1
    {’}{{'}}1
    {…}{{...}}1
    {≈}{{$\approx$}}1
    {×}{{$\times$}}1
    {≤}{{$\leq$}}1
    {≥}{{$\geq$}}1
}

\newcommand{\promptlisting}[2]{%
  \begin{tcolorbox}[
    enhanced,
    breakable,
    sharp corners,
    colback=white,
    colframe=black!45,
    colbacktitle=black!45,
    coltitle=white,
    fonttitle=\bfseries,
    title={#1},
    boxrule=0.35pt,
    left=1.5mm,
    right=1.5mm,
    top=1mm,
    bottom=1mm,
    before skip=0.85\baselineskip,
    after skip=0.85\baselineskip
  ]
  \lstinputlisting[style=promptbox]{#2}
  \end{tcolorbox}
}

\usepackage{xcolor}
\usepackage{xspace}

\definecolor{singlebg}{HTML}{C7DBF5}
\definecolor{singlefg}{HTML}{1B3A66}
\definecolor{divbg}{HTML}{E3CFF5}
\definecolor{divfg}{HTML}{4A2370}
\definecolor{anchorbg}{HTML}{C5EAD6}
\definecolor{anchorfg}{HTML}{14543A}
\definecolor{structbg}{HTML}{F7E0B8}
\definecolor{structfg}{HTML}{6B4012}

\newcommand{\cuebox}[3]{%
    {\textcolor{#1!35!#2}{\ttfamily\footnotesize\textbf{#3}}}\xspace}

\newcommand{\default}{\cuebox{singlebg}{singlefg}{vanilla}}
\newcommand{\diversified}{\cuebox{divbg}{divfg}{diversified}}
\newcommand{\persona}{\cuebox{anchorbg}{anchorfg}{position-guided}}
\newcommand{\defaultshort}{\cuebox{singlebg}{singlefg}{van.}}
\newcommand{\divshort}{\cuebox{divbg}{divfg}{div.}}

\title{Argument Collapse: LLMs Flatten Long-Form Public Debate}

\author{
Yekyung Kim\thanks{These authors contributed equally to this work.}
\quad
Yapei Chang\footnotemark[1]
\quad
Chau Minh Pham
\quad
Mohit Iyyer \\
University of Maryland, College Park \\
\texttt{\{yekyung,yapeic,chau,miyyer\}@umd.edu}
}

\begin{document}
\maketitle

\begin{abstract}
As LLMs are increasingly used to draft public-facing arguments, they may flatten public debate by repeatedly introducing the same polished, plausible arguments.
We study \emph{argument collapse}, the tendency of essays generated by different LLMs to converge to a smaller set of main arguments, sub-arguments, and paragraph-level structures.
We compare 1{,}039 human responses from 195 \textit{New York Times} (NYT) debates, 448 human responses from 61 longer-form \textit{Boston Review} (BR) forums, and 23{,}381 LLM-generated essays.
In the NYT corpus, 65.3\% of human main arguments are unique within a debate, compared to 3.4\% of LLM main arguments.
Asking LLMs to generate diverse answers adds variation, but a typical model recovers only about half of the distinct human main arguments, with much of the added variation falling outside the observed human argument space.
Collapse also appears in sub-arguments, where among essays with the same main argument, 41.0\% of human sub-arguments are unique versus 9.1\% from LLM responses.
Qualitatively, LLMs often reuse generalized and hedged sub-arguments, while humans prefer more concrete and topic-specific ones.
Structure-wise, LLM-generated essays tend to follow a more fixed arc, often opening with a direct claim and moving quickly toward proposals.
The same patterns hold in longer BR essays, suggesting that argument collapse extends beyond short-form responses. Finally, human-preference evaluators favor both common arguments and LLM essays, which could reinforce argument collapse.

\end{abstract}

\begin{center}
\small
\faGithub\ \href{https://github.com/mungg/argument_collapse}{\texttt{github.com/mungg/argument\_collapse}}
\end{center}

\section{Introduction} \label{sec:intro}

LLMs are now common aids in public-facing argumentative writing, including opinion essays and policy memos \citep{10.1145/3491102.3502030, Russell2025AIUI}. 
Such usage comes with a caveat: model suggestions can alter what claims writers make, how those claims are supported, and how much of the writer's own voice remains present~\citep{padmakumar2024doeswritinglanguagemodels, doshi-hauser-2024, abdulhai2026llmsdistortwrittenlanguage, röttger2026measuringmitigatingpersonadistortions}.
Prior work shows that LLMs can produce generative monocultures through narrowing output distributions \citep{wu2024generativemonoculturelargelanguage, zhang2025noveltybench, jiang2026artificial, nie2026perspectrascalableconfigurablepluralist} and reducing epistemic diversity in generated claims \citep{wright2026epistemicdiversityknowledgecollapse}. 
These homogenization measures, however, often do not directly compare human and LLM output distributions under the same task conditions, leaving it unclear what is lost when people write arguments with model assistance \citep{jain2026taskdependentevaluationllmoutput}.

In this work, we ask whether LLMs collapse into a narrower range of main arguments, supporting claims, and argumentative structures than human writers produce in response to the same debate questions. We use the term \emph{argument collapse} to describe cross-model convergence: the tendency of different LLMs, built by different frontier industry labs, to return to the same small set of plausible arguments rather than span the broader range of arguments humans make. Such failure has broad implications, as these systems can measurably shift reader beliefs \citep{jakesch-2023,fisher-2024}, narrow readers' perspectives \citep{sharma-2024,peterson-2024}, and recirculate through public discourse and training corpora in ways that amplify dominant arguments at the expense of long-tail reasoning \citep{shumailov2024curse}.

\begin{figure*}
    \centering
  \includegraphics[width=\linewidth]{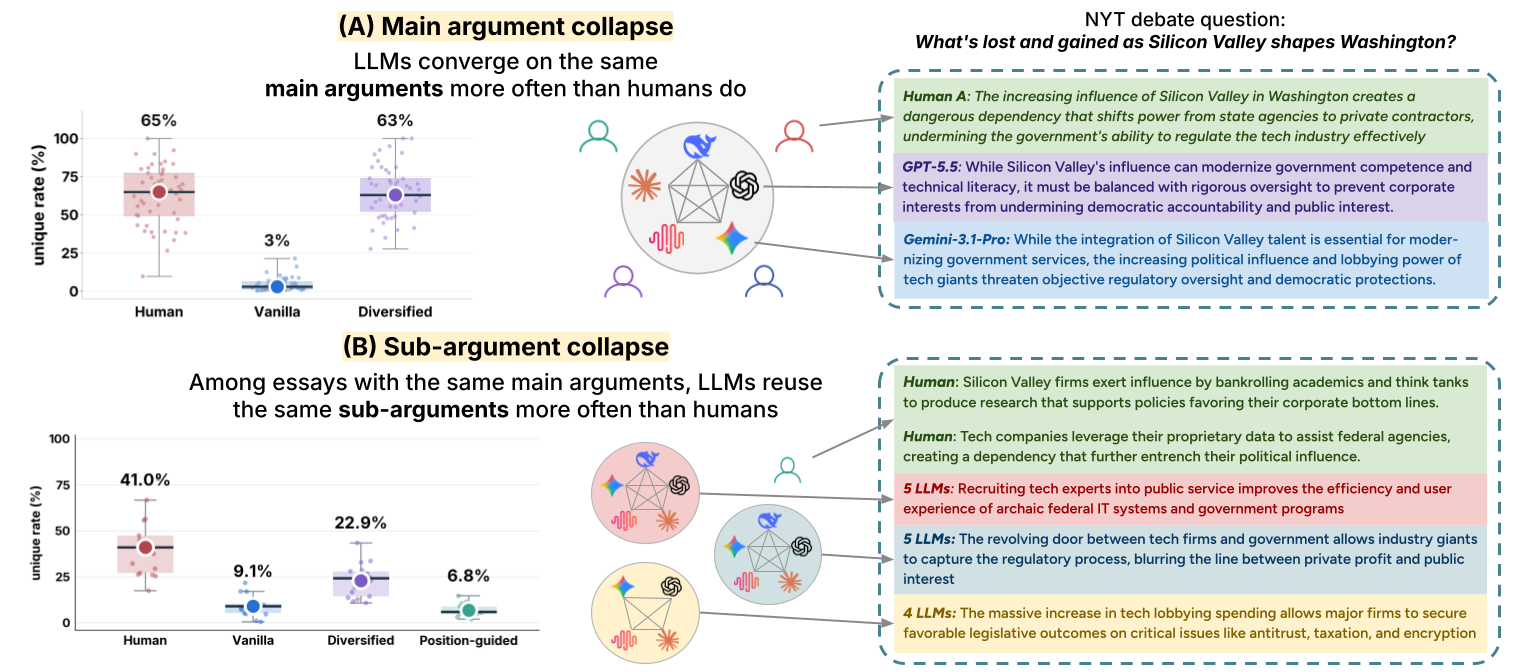}
  \caption{\textbf{Argument collapse at two levels of content.}
  \textbf{(A) Main-argument collapse:} LLMs converge on the same central arguments more often than human writers do.
  \textbf{(B) Sub-argument collapse:} among essays with the same main arguments, LLMs reuse the same supporting sub-arguments more often than human writers.}
    \label{fig:main-figure}
\end{figure*}

We study \emph{argument collapse} in three settings.
In the \default setting, we ask whether LLMs collapse even under a basic setup: when given a contested question, do their responses reflect the range of arguments humans make?
Because LLMs are often used for drafting and ideation \citep{Wan2023ItFL}, we test whether a \diversified prompt, which explicitly asks models to generate diverse responses, recovers the breadth of human arguments.
Finally, in the \persona setting, we provide the models with human responders' main argument, biography, and tone, then ask whether it can come up with the supporting reasons that the writer actually use.
We compare human-written responses with essays generated by five frontier LLMs across two corpora: \emph{New York Times Room for Debate} (NYT; $\approx$352 words) and longer \emph{Boston Review} forum responses (BR; $\approx$1{,}150 words). Our analysis focuses on main arguments, supporting reasons, and paragraph-level argumentative structure. Across all settings, we find evidence of \emph{argument collapse}:

\paragraph{Main arguments collapse under vanilla prompting.} Models from different providers repeatedly converge on the same main arguments (\autoref{fig:main-figure}A). In NYT, $65.3\%$ of human main arguments are unique within a debate, compared with only $3.4\%$ of \default LLM arguments. These arguments are often plausible and human-like, but are much less diverse than those produced by human writers.

\paragraph{Diversity prompting adds variation, but only partly recovers human arguments.} \diversified outputs are more varied than \default outputs, but they still miss many human-written arguments. A typical LLM recovers only about half of the distinct human main arguments ($50$--$55\%$), often missing narrower or more situated arguments.

\paragraph{Sub-argument collapse persists even when the main argument is fixed.} Among essays sharing the same main argument, only $9.1\%$ of \default sub-arguments are unique, compared with $41.0\%$ for humans (\autoref{fig:main-figure}B). \diversified and \persona outputs improve this only partially.

\paragraph{LLMs converge on different kinds of sub-arguments.} LLMs more often repeat generalized and hedged supporting arguments, whereas humans more often use concrete and topic-specific ones.

\paragraph{Collapse also appears in essay structure.} LLMs follow a more fixed structural arc. In NYT, \default LLMs move from \texttt{support} to \texttt{proposal} more than twice as often as humans ($29.4\%$ vs. $12.3\%$).

\section{Constructing a corpus of debates} \label{sec:data}

Measuring argument collapse requires multiple human and LLM responses to the same contested debate.
We use two public debate corpora with multiple responses to the same question.\footnote{We collect publicly accessible pages and linked debate pages from \url{https://www.nytimes.com/roomfordebate} and \url{https://www.bostonreview.net/forums/}.}

\subsection{Collecting human debate corpora}

\paragraph{NYT Room for Debate.}
Each NYT debate consists of one debate question and a set of invited human-written responses, each written as a short argumentative essay, with a median response length of 352 words. We collect the public-web archive, parse each page into a canonical layout with one question file and one markdown file per responder, and apply basic structural filters before creating the final corpus.
\footnote{Each debate must contain at least three responses, and each response must be at least 50 words.}

\paragraph{Boston Review.}
Each BR forum consists of a lead essay and commissioned replies.
This provides the same multiple-responses-to-one-debate structure as NYT, but in a longer form. Human responses have a median length of 1{,}150 words, with more room to develop their argument.

\paragraph{Filtering for durable debates.}
We exclude debates whose answers depend heavily on fast-changing events, because LLM responses to such debates would confound argument collapse with training-cutoff drift.
To do this, we tag each debate question for temporal change rate following the FreshQA framing \citep{vu2023freshllms}, and only keep debates where neither of two LLM taggers labels the debate question as \texttt{fast\_changing}.
\footnote{\texttt{fast\_changing} if the natural answer changes within roughly a year, \texttt{slow\_changing} if it changes over several years, or \texttt{never\_changing} if it is essentially static. We use \texttt{gpt-5.4-mini} and \texttt{gemini-3-flash-preview} as the two judges. They return a single label and a short rationale grounded in the debate question; model-call settings for both taggers are reported in \autoref{tab:model-hyperparameters}, and the tagging template is in \S\ref{app:prompts-preprocessing}. }

\paragraph{Final datasets.}
From the filtered NYT archive, we sample \textbf{195 debates} with broad topic coverage and a balanced question-type split: \textbf{1{,}039 human response essays} across 97 binary and 98 open-ended debates.
\footnote{We tag NYT debates by question type and broad subject area. Question-type tags identify binary debates with clear support/oppose sides for the stance analysis. Topic tags are used only to avoid a topically concentrated sample. The tagging templates are in \S\ref{app:prompts-preprocessing}.}
Our final BR corpus contains \textbf{448 human response essays} across \textbf{61 forums}. See \S\ref{app:artifact-use} for artifact-use and intended-use details.

\subsection{Collecting LLM responses}
For debates in both corpora, we generate corresponding responses from five frontier LLMs, GPT, Claude, Gemini, DeepSeek, and Minimax, under three generation conditions, \default, \diversified, and \persona.\footnote{Full model identifiers, model-call settings, and generation templates are reported in \autoref{tab:model-hyperparameters} and \S\ref{app:prompts-generation}.}
Across the three generation conditions, we collect \textbf{23{,}381 LLM essays}, with \textbf{16{,}661} for NYT and \textbf{6{,}720} for BR.\footnote{The NYT total decomposes as $5{,}195$ \default + $6{,}271$ \diversified + $5{,}195$ \persona essays. The \diversified count exceeds the naive $5\,\mathrm{models}\times 1{,}039\,\mathrm{humans}=5{,}195$ baseline because we additionally sample GPT under a higher reasoning-effort setting as an ablation. BR is $5\,\mathrm{models}\times 3\,\mathrm{conditions}\times 448\,\mathrm{humans}=6{,}720$.}

\paragraph{Vanilla.}
The \default condition measures each LLM's natural answer to the debate. We give the model only the debate question and sample one response per API call, repeating this $N$ times for each model, where $N$ is the number of human responders in the debate.\footnote{For BR, we just sample 5 times for each model per debate.}
From these samples, we identify one representative main argument for each model. The resulting comparison is between the human writers' main arguments and five model-level representative arguments, one per LLM.

\paragraph{Diversified.} The \diversified condition tests whether missing main-argument diversity can be elicited directly. We ask the model to produce a \emph{set} of $N$ essays in a single API call, where the instruction explicitly asks the model to vary central claims, supporting arguments, argumentative flow, and discourse moves as widely as possible.\footnote{Outputs are returned as a single text response with marker-delimited essays (\texttt{===== ESSAY N =====}) rather than JSON, because pilot runs showed that several models drop or truncate essays under structured-output constraints when asked for more than three at once.}

\paragraph{Position-guided.} We use \persona to test a stronger form of guidance. The LLM receives the debate question plus an anonymized sketch derived from one human responder in the same debate, which includes the human's main argument, bio, and tone. The LLM is then asked to write one essay from that writer's perspective.\footnote{Most instances in the NYT dataset already come with an author bio. To reduce the risk that LLMs recall specific humans from training data, we anonymize this bio with \texttt{gemini-3-flash-preview}. Whenever an author bio is not present, we ask the judge to derive a rough sketch based on the human essay itself. See details in \S\ref{app:prompts-preprocessing}.}

\subsection{Corpus and stance annotations} \label{sec:stance-annotations}

Using \texttt{gemini-3-flash-preview}, we tag each NYT debate by question type, distinguishing binary debates with clear support/oppose sides from open-ended debates.
This yields 97 binary and 98 open-ended debates. For binary debates, we also label each essay's final stance. A judge first
identifies the debate's support and oppose sides as concise statements, and a second judge labels each essay as
\texttt{strong\_oppose}, \texttt{weak\_oppose}, \texttt{neutral}, \texttt{weak\_support}, or \texttt{strong\_support} relative to
those sides.
One author independently annotate a random sample of 20 essays using the same stance schema, reaching 70\% agreement with the judge's annotations ($\kappa=0.625$).

\begin{table*}[t]
    \centering
    \small
    \newcommand{\argcountcell}[2]{\makebox[3.1em][c]{\colorbox{#1}{\strut\hspace{2pt}#2\hspace{2pt}}}}
    \renewcommand{\arraystretch}{1.22}
    \setlength{\tabcolsep}{4.5pt}
    \begin{tabularx}{\textwidth}{@{}X
        >{\centering\arraybackslash}p{0.075\textwidth}
        >{\centering\arraybackslash}p{0.085\textwidth}
        >{\centering\arraybackslash}p{0.095\textwidth}@{}}
    \toprule
    \textbf{Representative main argument from the ``Are Americans Too Obsessed With Cleanliness?'' debate} &
    \makecell{\textbf{Humans}\\[-0.1em]{\scriptsize writers}} &
    \makecell{\textbf{\textcolor{vancolor}{\defaultshort}}\\[-0.1em]{\scriptsize reps}} &
    \makecell{\textbf{\textcolor{divcolor}{\divshort}}\\[-0.1em]{\scriptsize families}} \\
    \midrule
    \addlinespace[0.25em]
    \multicolumn{4}{@{}l}{\textbf{\textcolor{secfg}{(1)\,\,\default LLM responses converge on one hedged argument}}}\\
    \textit{While Americans should maintain essential hygiene practices that prevent disease, they should abandon the neurotic pursuit of total sterility driven by social anxiety and marketing.} &
    \argcountcell{singlebg!75}{$2/9$} &
    \argcountcell{singlefg!28!singlebg}{\textbf{$5/5$}} &
    \argcountcell{singlefg!28!singlebg}{\textbf{$5/5$}} \\
    \midrule
    \addlinespace[0.25em]
    \multicolumn{4}{@{}l}{\textbf{\textcolor{secfg}{(2)\,\,\diversified recovers additional human arguments}}}\\
    \textit{The United States must prioritize and improve hygiene education and enforcement, specifically hand washing, to prevent illness and save lives.} &
    \argcountcell{singlebg!55}{$1/9$} &
    \argcountcell{singlebg!18}{$0/5$} &
    \argcountcell{singlebg!95}{$4/5$} \\
    \textit{American obsession with cleanliness was initiated by 19th-century social factors and is sustained by advertising campaigns that exploit social anxieties.} &
    \argcountcell{singlebg!55}{$1/9$} &
    \argcountcell{singlebg!18}{$0/5$} &
    \argcountcell{singlebg!95}{$4/5$} \\
    \midrule
    \addlinespace[0.25em]
    \multicolumn{4}{@{}l}{\textbf{\textcolor{secfg}{(3)\,\,\diversified rarely recovers more distinctive human arguments}}}\\
    \textit{Hygiene practices and their definitions are culturally relative rather than universal, frequently leading to misunderstandings between different societies.} &
    \argcountcell{singlebg!55}{$1/9$} &
    \argcountcell{singlebg!18}{$0/5$} &
    \argcountcell{singlebg!55}{$1/5$} \\
    \textit{Rigid control and rule-following in suburban life are ultimately futile because true grace and peace come from accepting life's inherent messiness.} &
    \argcountcell{singlebg!55}{$1/9$} &
    \argcountcell{singlebg!18}{$0/5$} &
    \argcountcell{singlebg!55}{$1/5$} \\
    \textit{Religious purification rituals primarily facilitate a search for the sacred and must be understood beyond mere psychological wellness or neurosis.} &
    \argcountcell{singlebg!55}{$1/9$} &
    \argcountcell{singlebg!18}{$0/5$} &
    \argcountcell{singlebg!55}{$1/5$} \\
    \midrule
    \addlinespace[0.25em]
    \multicolumn{4}{@{}l}{\textbf{\textcolor{secfg}{(4)\,\,\diversified introduces arguments no human raised}}}\\
    \textit{Americans should relax their extreme standards of cleanliness because our obsession with sterilization isolates us and creates an unnecessary barrier to authentic human connection.} &
    \argcountcell{singlebg!18}{$0/9$} &
    \argcountcell{singlebg!18}{$0/5$} &
    \argcountcell{singlebg!55}{$1/5$} \\
    \textit{Americans are not too obsessed with cleanliness; rather, they are dangerously inconsistent in their hygiene practices when moving from private to public spaces.} &
    \argcountcell{singlebg!18}{$0/9$} &
    \argcountcell{singlebg!18}{$0/5$} &
    \argcountcell{singlebg!55}{$1/5$} \\
    \bottomrule
    \end{tabularx}
    \caption{\textbf{Main-argument collapse and partial recovery in one cleanliness debate.} This debate asks whether Americans are too obsessed with cleanliness. Rows show representative main arguments from the observed overlap patterns. Counts indicate how many human writers, \default LLM representatives, or \diversified LLM families produced a substantially overlapping argument; for \diversified, a family counts once if any of its diversified answers matches the row. All five \default representatives converge on the same hedged argument. \diversified generation recovers some additional human arguments, rarely reaches more distinctive human arguments, and introduces arguments not raised by any human writer. See more examples in \S\ref{app:main-arg-recovery-types}.}
    \label{tab:main-arg-recovery-examples}
    \vspace{-8pt}
\end{table*}

\section{LLM collapse at the content level} \label{sec:content}

The paired human and LLM responses let us measure collapse in the substance of the arguments themselves. Content collapse occurs when separately generated essays return to the same arguments rather than spreading across the range of arguments human writers make. We investigate this phenomenon for \textit{main arguments} and \textit{sub-arguments}, finding collapse at both levels.

\subsection{Analysis setup}

We use one pipeline for both main arguments and sub-arguments to extract argument units from each essay, then label pairwise overlap between same-debate units.
All extraction and labeling steps in this pipeline use \texttt{gemini-3-flash-preview}.
See all prompts in \S\ref{app:prompts-content}.\footnote{Model-call settings for extraction and labeling are reported in \autoref{tab:model-hyperparameters}.}

\paragraph{Argument extraction.} An essay's main argument is the overall claim it defends, and a sub-argument is a discrete supporting claim, piece of evidence, warrant, or qualification that develops or backs the main argument \citep{toulmin1958uses,stab-gurevych-2017-parsing}.
See example main and sub-arguments in \autoref{fig:main-figure}.
We extract one main argument and a list of sub-arguments per essay using an LLM judge.\footnote{One author validates a random sample of $30$ human essays for both precision and recall: all $30$ extracted main arguments matched the essay, all $134$ extracted sub-arguments were present in the source text, and no substantive supporting sub-arguments were missing.}


\paragraph{Labeling pairwise argument overlap.}
For each pair of same-type argument units within the same debate,\footnote{That is, either a pair of main arguments or a
pair of sub-arguments.} we ask the judge how much the two units overlap using one of four labels: \texttt{equivalent}, \texttt{strong\_overlap}, \texttt{weak\_overlap}, and \texttt{different}.
Two author annotators manually label 100 same-debate main-argument pairs to validate this schema.
The two annotators reach $\kappa=0.61$ on exact four-label agreement and $\kappa=0.80$ after collapsing labels into substantially overlapping (\texttt{equivalent} or \texttt{strong\_overlap}) versus not substantially overlapping (\texttt{weak\_overlap} or \texttt{different}).
\footnote{The final judge agreed with the two annotators on 69\% and 63\% of exact labels, and on 93\% of coarse labels for both annotators. See validation details and example pairs in \S\ref{app:pairwise-validation}.}
Unless specified otherwise, we treat \texttt{equivalent} and \texttt{strong\_overlap} pairs as \textit{substantially overlapping}.

\paragraph{Unique rate.}
For both main and sub-arguments, we ask how often an argument unit is genuinely unique within the group being compared.\footnote{A comparison group is the set being evaluated for internal repetition within the same debate, such as human-written arguments or LLM-generated arguments.} An argument is reused if it has a substantial overlap with another argument from the same debate and group; otherwise it is \emph{unique}. Because groups can contain different numbers of arguments, we compare them at the same sample size: if we drew $m$ arguments from each group within a debate, what fraction would have no substantial-overlap match inside that sample? Formally, for a group $G$, let $d_i$ be the number of substantial-overlap matches for argument $i \in G$. At sample size $m$,
\[
U_m(G)=\frac{1}{|G|}\sum_{i\in G}
\frac{\binom{|G|-1-d_i}{m-1}}{\binom{|G|-1}{m-1}}.
\]
This is the expected fraction of sampled arguments with no match inside the sample. When $m=|G|$, this is simply the fraction of all arguments in the group that have no match. Higher values mean more distinctive arguments; lower values mean more within-group reuse.
\footnote{For main arguments, each essay contributes one unit. For sub-arguments, an essay can contribute several supporting reasons, so the reported rate is a share of extracted sub-arguments rather than a share of essays.}
Other recovery and reuse metrics used in the content analyses are summarized in \autoref{tab:content-metrics-details}.


\subsection{Main argument collapse across vanilla and diversified settings} \label{sec:main_arg}

We ask how models, under different generation settings, vary in the main arguments they produce.

\subsubsection{LLMs naturally collapse under vanilla prompting}

When models receive only the contested debate question, they consistently repeat the same main argument across different debate types. These repeated arguments often overlap with at least one human argument in the same debate, but LLM essays also show weaker stance strength.\footnote{To keep group sizes comparable, we report $U_m$ with $m = \min(N_{\text{human}}, 5)$.} 


\paragraph{Vanilla responses repeat human-like arguments.}
LLMs from different providers repeatedly converge on the same main arguments, while human writers more often introduce unique claims. As shown in \autoref{fig:main-figure}A, across 195 NYT debates, $65.3\%$ of human main arguments are unique within a debate, compared with only $3.4\%$ of \default LLM arguments. The same pattern holds in the longer BR essays ($78.6\%$ vs. $18.4\%$), with the human rate higher in $58$ of $61$ forums. The repeated LLM arguments usually remain within the human argument space. In NYT, $77\%$ of \default LLM main arguments substantially overlap with at least one human main argument from the same debate. For example, in \autoref{tab:main-arg-recovery-examples}, all five LLMs produce the  similar hedged \textit{``concede basic hygiene, prescribe moderation''} argument, while human writers also raise several other main arguments.

\paragraph{On binary debates, LLMs show weaker stance strength.}
For debates with clearly defined support and oppose sides,\footnote{Identified and labeled using the stance-annotation procedure in \autoref{sec:stance-annotations}.} we evaluate both side balance and stance strength.
Counting both weak and strong labels, \default LLM essays support the proposition somewhat more often than human essays ($56.1\%$ vs. $49.7\%$) and oppose it less often ($34.5\%$ vs. $43.1\%$); the remaining essays are neutral or noncommittal ($9.4\%$ vs. $7.2\%$).
The sharper gap is stance strength: $76.1\%$ of human essays take a strong support or oppose stance, compared with $63.4\%$ of \default LLM essays.

\subsubsection{LLM behavior under diversified prompting}

The \default results show that models converge when asked for a single natural answer. We next test whether collapse persists when models are explicitly asked to produce several diverse answers, as users might do when brainstorming.

\paragraph{Diversified prompting increases main argument uniqueness.}
\diversified prompting increases the overall unique rate, though the improvement varies by model. GPT remains below the human baseline: $45\%$ of its diversified main arguments are unique, compared with $65.3\%$ of human main arguments in the same debates.
Minimax ($53\%$) and Claude ($58\%$) are closer to the human rate, DeepSeek is roughly human-level ($63\%$), and Gemini exceeds the human unique rate ($82\%$).

\paragraph{Diversified prompting only partly recovers human argument diversity.}
\diversified prompting increases uniqueness, but it does not fully recover the range of arguments human writers make: a typical \diversified LLM covers only about half of human main-argument clusters, ranging from $50\%$ for Claude to $55\%$ for Gemini. Pooling all five models raises coverage to $73.9\%$, but mostly by recovering easier-to-find arguments: arguments made by multiple human writers are recovered $98.1\%$ of the time, compared with $67.8\%$ for one-off human arguments. BR shows the same partial recovery pattern: pooling five \diversified models recovers $63.1\%$ of human main-argument clusters.

\paragraph{Many diversified arguments have no observed human counterpart.}
From the LLM side, many \diversified arguments have no observed human counterpart: only $47.6\%$ to $60.3\%$ of NYT main arguments, and only $27.6\%$ of pooled BR main arguments, substantially overlap with something humans actually said.
Broad and direct answers are easier to recover, while specific and narrower proposals are more often missed.
The cleanliness debate in \autoref{tab:main-arg-recovery-examples} gives a concrete example: LLMs recover the hand-washing-as-vital and cultural-construct arguments, miss the culturally-relative, control-over-chaos, and religious-impulse arguments, and add an authentic-human-connection argument not observed among the human writers (additional examples in \S\ref{app:main-arg-recovery-types}).

\paragraph{Diversified prompting balances sides, but not stance strength.}
For binary debates, \diversified prompting makes LLMs look more like humans in which side they choose. The share of LLM responses supporting the debate proposition falls from $56.1\%$ under \default to $49.6\%$ under \diversified, nearly identical to the human share of $49.7\%$.
But only $66.4\%$ of \diversified responses take a strong support or strong oppose stance, compared with $76.1\%$ of human responses.

\paragraph{Collapse is not explained by memorization.}
To isolate the effect of potential memorization on argument collapse, we focus on the six most recent BR forums\footnote{We exclude NYT, as Room for Debate ended in 2018.} (May 2025 to February 2026) with post-cutoff forums for three of five models. The pattern persists with human arguments remaining more unique than \default arguments at $84.0\%$ compared with $35.6\%$, while \diversified prompting recovers only $53.8\%$ of human main arguments (\S\ref{app:memorization}).

\subsection{Sub-argument collapse} \label{sec:sub_arg}
Having established main-argument collapse, we next ask whether the collapse appears in supporting arguments. This question is important because even when essays share the same main argument, they can develop it through different lines of support.
To isolate sub-argument collapse, we focus on debates where humans and LLMs have shared main arguments.\footnote{We filter the cohorts to ensure that at least three humans share the same main argument and that, on the LLM side, all five LLMs have at least one essay aligned with that same argument. Full details in \S\ref{app:shared-main-arg-sampling}.} This subset contains $16$ NYT debates, $62$ human essays, $80$ \default essays, $321$ \diversified essays and $310$ \persona essays.
Within each cohort, we compute $U_m$ for humans and three LLM setups using a common sample size $m$ , then macro-average across cohorts.\footnote{$m = \min(|H|, |V|, |D|, |P|)$ where $H,V,D,P$ are the humans, \default, \diversified, \persona
sub-argument pools, so all four groups are compared at the same sample size within each cohort.} 

\begin{figure}[t]
\centering
\includegraphics[width=\linewidth]{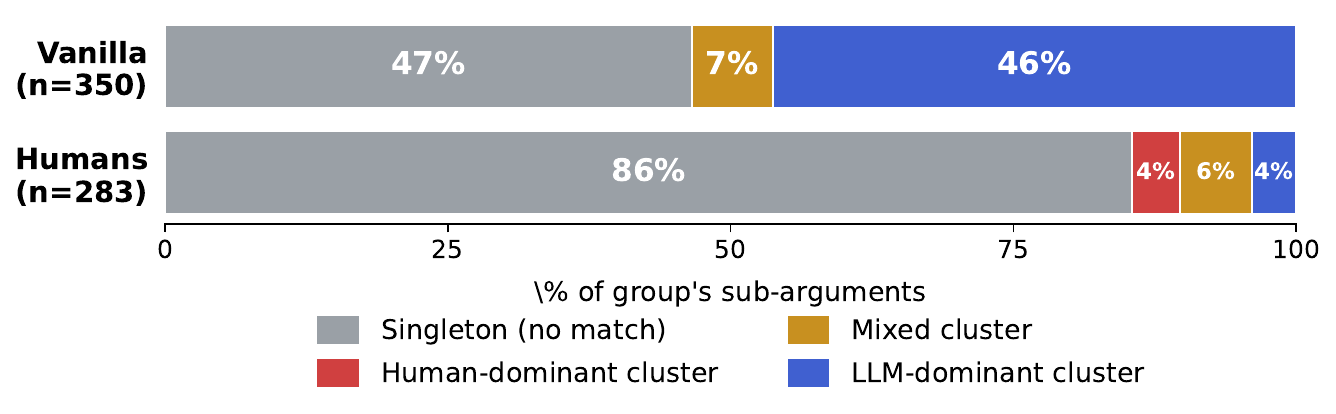}
\caption{\textbf{Per-group distribution of sub-arguments.} Each bar shows the share of a group's sub-arguments in singleton clusters or multi-member clusters with $\geq 70\%$ human or LLM members. Details in \S\ref{app:cluster-ratio-distribution}.}
\label{fig:cluster-ratio-overview}
\vspace{-15pt}
\end{figure}

\begin{table*}[t]
\centering
\footnotesize
\renewcommand{\arraystretch}{1.08}
\setlength{\tabcolsep}{3pt}
\begin{tabularx}{\textwidth}{@{}p{0.17\textwidth}X@{}}
\toprule
  \multicolumn{2}{@{}c@{}}{\cellcolor{red!18}\textbf{Human examples}} \\
\midrule
\emph{Specific case}
& ``The case of \textbf{Hurricane Sandy} demonstrates that \textbf{inadequate flood-protection studies and infrastructure investment}, ...'' \\

\emph{Specific cause-effect}
& ``Heavy-handed federal investment in drug enforcement has led to \textbf{over half of the federal prison population being incarcerated for drug offenses}.''  \\

\emph{Concrete solution}
& ``\ldots could be prevented if companies implemented simple, low-cost measures like \textbf{complex passwords and frequent software patching}.'' \\[-1pt]
& ``Building \textbf{bicycle lanes that are physically separated from car traffic} is essential for ensuring the safety of all demographics, including children.''  \\
\midrule
  \multicolumn{2}{@{}c@{}}{\cellcolor{blue!18}\textbf{LLM examples}} \\
\midrule
\emph{Generic citation}
& ``\textbf{Research indicates} that sex offenders have lower recidivism rates than other felons and that residency restrictions do not actually reduce reoffending.'' {\scriptsize (by Claude)} \\

\emph{Abstract concept}
& ``Cybersecurity failures create significant \textbf{externalities where the damage of a breach extends far beyond the responsible company} to the general public.'' {\scriptsize (by Minimax)} \\

\emph{Abstract solution}
& ``Crime laboratories should be \textbf{independent from police and prosecutors} to insulate analysts from investigative pressure and cognitive bias.'' {\scriptsize (by GPT)} \\

\emph{Hedged generality}
& ``A \textbf{global framework must guarantee} surrogates \textbf{comprehensive medical care and long-term health insurance} to protect them from physical complications.'' {\scriptsize (by Gemini)} \\[-1pt]
& ``\textbf{Effective maternal care requires investing in a continuum of support}, such as home-visiting programs and practical assistance, rather than ...'' {\scriptsize (by DeepSeek)} \\
\bottomrule
\end{tabularx}

\vspace{-4pt}
\caption{\textbf{Example sub-arguments from humans and LLMs in NYT debates.} Italic tags describe the anchor used.}
\label{tab:human-llm-features}
\vspace{-10pt}
\end{table*}

\paragraph{Sub-argument collapse persists even under shared main arguments.}
Only $9.1\%$ of \default sub-arguments are unique, compared with $41.0\%$ for humans. \diversified recovers some diversity but reaches only $22.9\%$ in \autoref{fig:main-figure}B.
The \persona condition asks whether grounding LLMs in human perspectives restores more varied supporting reasons. As a sanity check, we first hold the target writer fixed and compare the five LLMs conditioned on that same writer.
Their sub-arguments become highly similar ($6.8\%$ unique rate), confirming that position guidance does anchor models to the assigned perspective.
We then ask the more substantive question: if the assigned writer changes, will the same LLM produce a wider range of sub-arguments?
Holding the LLM fixed and varying the target writer raises unique rate to $18.4\%$ on average across the five LLMs, still well below the human rate.
The most diverse LLMs are Minimax and Claude, but they reach only $22.2\%$ and $21.4\%$.
See more details in \S\ref{app:shared-main-arg-sampling}.

\paragraph{LLMs converge on generalized arguments and humans on concrete, topic-specific ones.}
The distribution of sub-argument clusters suggests a clear difference between humans and \default LLMs (\autoref{fig:cluster-ratio-overview}). Humans have $85.5\%$ of their sub-arguments in singleton clusters, whereas \default LLMs have only $46.6\%$ in singletons and another $46.3\%$ in clusters where at least $70\%$ of
sub-arguments come from LLMs (\emph{LLM-dominant} clusters). To understand these qualitative differences, we compare sub-arguments from LLM-dominant against human-dominant clusters, and each singletons (details in \S\ref{app:cluster-ratio-qualitative}). We observe that arguments from humans often stay closer to specific instances. Humans tend to ground their arguments in a specific case, or a concrete intervention and they take a direct position with less hedging.
In comparison, LLM arguments more often reach for generalized frameworks such as abstract concepts, cite generic appeals rather than naming a specific case, abstract institutional interventions and hedged generalities that balance multiple stakeholders without committing to any specific action. \autoref{tab:human-llm-features} shows the examples. LLM collapse is not just repeated content, but convergence onto generalized frameworks that are hard to falsify and hedged framings that stay on safe ground. One possible explanation is that alignment favor safer and more broadly acceptable responses, pushing models toward generalized supports \citep{zhang2025verbalizedsamplingmitigatemode, yun2025priceformatdiversitycollapse, kirk2024understandingeffectsrlhfllm}.

\paragraph{Sub-argument collapse persists in longer essays.}
To test whether sub-argument collapse generalizes in longer essays, we apply the within-group unique rate analysis from \S\ref{sec:sub_arg} to a Boston Review subset of $16$ forums, 60 human responses and 70 \default essays.\footnote{We apply less strict filtering to BR because it has a smaller dataset and more fine-grained main arguments, so only a few cohorts satisfy the filter. See \S\ref{app:BR-shared-main-arg-sampling}.}
For each forum, we identify each writer group's largest main-argument cluster and compute $U_m$ separately
within each, then macro-average across forums. Across the $16$ qualifying forums, the unique rate is $16.3\%$
for \default LLMs and $42.2\%$ for humans. This result aligns with what we observed in NYT despite the longer essays and different format. See more details in \S\ref{app:BR-shared-main-arg-sampling}.

\section{LLM collapse at the structure level} \label{sec:structure}

The same argument content can be built in many ways: a writer might open with a direct claim, develop evidence slowly, concede an opposing view, narrate an example, or end with a policy proposal. We therefore ask whether LLM collapse also appears at the structural level. The answer is yes: LLM essays are more likely to follow the same structural arc, opening with a direct claim, moving through support, and ending with a proposal, while human essays vary more in how they build and develop the argument.

\subsection{Measuring structural collapse}

The structure analysis asks whether essays are organized in similar ways, regardless of the specific arguments they make. We compare paragraph-label patterns directly.

\paragraph{Paragraph-level annotation.} For each essay, we tag every paragraph along two orthogonal layers. The \emph{argumentative-role} layer assigns the paragraph's role in the essay's progression: \texttt{thesis}, \texttt{support}, \texttt{counterclaim}, \texttt{rebuttal}, \texttt{concession}, \texttt{reframing}, \texttt{implication}, \texttt{proposal}, or \texttt{none} (full definitions in \S\ref{app:structure-taxonomies}). The \emph{discourse-mode} layer captures how the paragraph is written, independent of its argumentative role: \texttt{argumentation}, \texttt{exposition}, \texttt{narration}, or \texttt{description}. The two layers run as separate annotation passes with the same configuration.\footnote{Each annotation pass receives the full essay text and the layer-specific taxonomy; model-call settings are reported in \autoref{tab:model-hyperparameters},annotation prompts are in \S\ref{app:prompts-structure}, and human validation results are in \autoref{tab:annotation-validation}.}

\paragraph{Structural summaries.} We summarize structure in two ways. First, we divide each essay into eight normalized position bins and measure which paragraph labels appear near the beginning, middle, and end. Second, we measure paragraph-to-paragraph transitions, such as whether a \texttt{support} paragraph is followed by more \texttt{support} or by a \texttt{proposal}.

\subsection{Evidence for structural collapse}

\paragraph{LLMs follow a more fixed structural arc.}
Human essays have a recognizable but flexible organization: they need not begin with a compact \texttt{thesis}, and they continue mixing support, explanation, and occasional narration across the essay. LLM essays are more regular. In both NYT and BR, \default LLM responses are more likely than humans to open with \texttt{thesis}, move into \texttt{support}, and end with \texttt{proposal}. In terms of discourse mode, LLM paragraphs are more likely to be labeled \texttt{argumentation}, while human responses tend to mix in more \texttt{exposition} throughout. The same pattern appears in guided generation, as shown in \autoref{fig:structure-aggregate-heatmap} and \autoref{fig:structure-aggregate-heatmap-br}.

\paragraph{Humans develop support, LLMs move sooner to proposals.}
In terms of paragraph transitions, human essays are more likely to continue developing support. \texttt{support} $\rightarrow$ \texttt{support} appears in $50.5\%$ of NYT and $54.5\%$ of BR human transitions, compared with $36.0\%$ and $29.7\%$ for \default responses.
\default LLMs instead move from support to resolution more often: \texttt{support} $\rightarrow$ \texttt{proposal} appears in $29.4\%$ of NYT LLM support transitions and $17.7\%$ of BR LLM support transitions, compared with $12.3\%$ and $7.2\%$ for humans.
See details in \autoref{tab:structure-flow-patterns} and \autoref{tab:structure-flow-patterns-br}.


\paragraph{Diversified and position-guided essays still look structurally LLM-like.}
\diversified and \persona essays remain close to the \default arc. In NYT, they slightly reduce the tendency to open with \texttt{thesis} and slightly increase \texttt{exposition} or \texttt{narration} discourse, but \texttt{argumentation} still accounts for $89.7\%$ of \diversified paragraphs and $89.1\%$ of \persona paragraphs, compared with $71.5\%$ for humans. In BR, the guided conditions also move some paragraph-position trends toward humans, but generated essays still rely more heavily on \texttt{argumentation} and show less variation than human responses.

\section{Do Evaluators Reward Collapse?}
\label{sec:evaluators}

As discussed in \S\ref{sec:sub_arg}, one possible explanation for argument collapse is that alignment favors broadly acceptable responses over rare ones. We therefore ask whether widely-used evaluators show a similar preference. We examine two reward models trained on human preference data, Skywork-Reward-V2 \citep{liu2025skywork} and ArmoRM \citep{wang2024armorm}, two LLM-as-a-judge protocols following MT-Bench \citep{zheng2023judging} and WildBench \citep{lin2025wildbench}, and two quality-based evaluators, an LLM judge scoring \emph{argument quality} with the 15 dimensions of \citet{wachsmuth-2017} and a creative-writing preference reward model (\emph{Writing RM}; \citealp{fein2026litbench}). See more details in \S\ref{app:evaluator-details}.

\paragraph{Evaluators prefer common arguments over rare ones.}
We compare shared- against unique-argument essays within the $102$ debates where both types occur in the human and \diversified{} LLM pools (\autoref{tab:evaluators}, left). All six evaluators favor shared-argument essays, with all but one comparison significant. Human-preference evaluators show a clear preference for shared arguments, perhaps reflecting a tendency to favor more widely held views.\footnote{This preference holds even when scoring the main argument level (\S\ref{app:evaluator-robustness}).} The effect is weaker but persists under evaluations of argument quality and creative writing quality. This result highlights a potential incentive for evaluators to reward more common arguments.

\paragraph{Evaluators prefer the LLM essay under same argument.}

We compare $550$ human and \default{} pairs across $140$ debates with equivalent main arguments, evaluating both the full essays and main arguments (\autoref{tab:evaluators}, right). All evaluators except \emph{Writing RM} favor LLM outputs at both levels, with the preference particularly strong for full essays. The preference persists after paraphrasing human claims (\S\ref{app:evaluator-robustness}), indicating a potential incentive for how LLMs develop the same argument.

\begin{table}[t]
  \centering\footnotesize
  \renewcommand{\arraystretch}{1.12}\setlength{\tabcolsep}{3pt}
  \begin{tabular}{@{}lcccc@{}}
  \toprule
  \multirow{3}{*}{\textbf{Evaluator}} & \multicolumn{2}{c}{\textbf{Shared vs.\ unique:}} & \multicolumn{2}{c}{\textbf{Same argument:}} \\
  & \multicolumn{2}{c}{\textbf{shared wins (\%)}} & \multicolumn{2}{c}{\textbf{LLM wins (\%)}} \\
  \cmidrule(lr){2-3}\cmidrule(l){4-5}
  & \textbf{Human} & \textbf{LLM} & \textbf{Main Arg} & \textbf{Essay} \\
  \midrule
  Skywork-V2 & 74.7 & 72.3 & 84.3 & 100.0 \\
  ArmoRM & 67.0 & 66.3 & 77.5 & 100.0 \\
MT-Bench & 64.3 & 69.4 & 86.9 & 89.3 \\
WildBench & 70.3 & 64.0 & -- & 89.5 \\
  Argument quality & 62.0 & 63.7 & 62.7 & 96.2 \\
  Writing RM & 59.3\rlap{$^{\dagger}$} & 61.8 & 55.0\rlap{$^{\dagger}$} & 50.7\rlap{$^{\dagger}$} \\
  \bottomrule
  \end{tabular}
\caption{\textbf{Evaluator preferences on shared-vs.-unique comparisons and same-argument pairs.} Each number is the percentage of debates favoring the shared argument (left) or the LLM essay (right) under debate-level sign tests ($\dagger$: not significant; all others $p<0.05$). WildBench is essay-only due to its checklist-based criteria.}

  \label{tab:evaluators}
  \end{table}

\section{Related Work} \label{sec:related_work}

\paragraph{Diversity collapse and surface idiosyncrasy in LLM writing.}
\emph{Model collapse} describes tail loss when models are recursively trained on their own outputs \citep{shumailov2024curse}. Prior work finds lexical and content diversity loss in co-writing and revision \citep{padmakumar2024doeswritinglanguagemodels, anderson-2024, abdulhai2026llmsdistortwrittenlanguage}, semantic and voice shifts under minimal editing \citep{jiang2026artificial, röttger2026measuringmitigatingpersonadistortions}, and constrained coverage of political and epistemic content \citep{santurkar-etal-2023, argyle-etal-2023, röttger2025issuebenchmillionsrealisticprompts, wright2026epistemicdiversityknowledgecollapse, durmus2024measuringrepresentationsubjectiveglobal}. Other work attributes these patterns to typicality bias, prompt format, or alignment pressures \citep{zhang2025verbalizedsamplingmitigatemode, yun2025priceformatdiversitycollapse, chakrabarty2025aislop, tu2026sharednatureuniquenurture}.
Prior work shows models generate recognizable lexical and stylistic patterns even under paraphrase or style changes \citep{sun-2025-idiosyncrasies,
bitton-2025-fingerprints, russell2026storyscopeinvestigatingidiosyncrasiesai}. 
LLM-assisted writing shifts user beliefs, narrows brainstorming, and influences
decisions such as voting in field settings \citep{jakesch-2023, sharma-2024, fisher-2024}, and
\citet{wen2026automatedw2s} introduce \emph{entropy collapse} in automated research idea generation, where ideas converge on a narrow shared set across LLMs and prompts. In this work, we study longer form essays and measure diversity across arguments, supporting reasons, and rhetorical structure.

\paragraph{Argument-level evaluation and mitigation.}
Argumentative content has long been studied through argument mining \citep{stab-gurevych-2017-parsing, wachsmuth-2017, gupta-etal-2024-harnessing}, discourse analysis \citep{hyland-2005} and quality assessment comparing with human and llm \citep{Herbold2023ALC}. More recent work asks how to recover lost diversity in
LLM outputs through prompting or sampling
\citep{zhang2025verbalizedsamplingmitigatemode, hayati-etal-2024-far}, configurable pluralism prompts
\citep{nie2026perspectrascalableconfigurablepluralist}, multi-perspective generation
\citep{wu2024generativemonoculturelargelanguage, zhang2025noveltybench}, and lightweight editing pipelines
\citep{jiang2026artificial}. Persona prompting goes further by conditioning generations on a target identity,
role, or stance \citep{samuel-2024-personagym, li-2025-persona-promise, du-2025-twinvoice,
shin-etal-2025-spotting}. Related work studies strategy-level patterns in LLM-generated conversation \citep{poungpeth2026spontaneouspersuasionauditmodel}.

\section{Conclusion}
We define \emph{argument collapse} as the tendency of independently generated essays to converge on the same small set of plausible arguments. Across \textit{New York Times} debates and longer-form \textit{Boston Review} forums, we find collapse at three levels: main arguments, supporting reasons, and argumentative structure. Model-generated essays converge on fewer main arguments, reuse supporting reasons more often even under a shared main argument, and follow a more standardized argumentative arc.  Moreover, automatic evaluators reward shared arguments more than unique ones. The risk is not only surface idiosyncrasy or opinion bias, but a narrowing of the range of arguments readers encounter, potentially amplifying dominant ones and limiting long-tail reasoning.
\section*{Limitations}

Our study focuses on measuring argumentative diversity rather than argument quality or factual accuracy. Even if humans generate more distinctive arguments, this does not necessarily mean those arguments are better than others. Thus, our results should not be interpreted as showing that human essays are always more persuasive, more accurate, or more preferable than LLM-generated essays.
Second, the essays were written at different times. The human essays were written earlier, while LLM responses were generated later and may reflect differences in training data and temporal knowledge. We try to address this concern by filtering out debates that depend heavily on fast-changing events. However, subtle temporal mismatches cannot be fully eliminated, so this remains a limitation. Some of the older debates may also have appeared in the models' training data. We find the same collapse pattern in recent Boston Review forums published after the models' training cutoffs (\S\ref{app:memorization}), but cannot fully rule out memorization effects for older debates.
Third, our dataset focuses on public debate forums, so our findings may not transfer directly to other domains, such as research writing or legal reasoning.
Lastly, several parts of our analysis rely on LLM-based annotations. While we report inter-annotator agreement (IAA), these annotations are still imperfect and may introduce systematic biases.

\section*{Ethical considerations}
LLMs are used for writing assistance, not for generation from scratch. 
\section*{Acknowledgments}

We thank the University of Maryland Computational Linguistics and Information Processing (CLIP) Lab for their feedback and support. This project was partially supported by awards IIS-2626013 and IIS-2545884 from the National Science Foundation (NSF). We also thank Google for a Cloud Credit award that enabled this research. We also thank Professor Shi Feng for feedback on an early version of the work.


\bibliography{custom}

\appendix

\section{Data Appendix}
\label{app:data}

\begin{table*}[!t]
\centering
\small
\renewcommand{\arraystretch}{1.12}
\setlength{\tabcolsep}{5pt}
\begin{tabular*}{\textwidth}{@{\extracolsep{\fill}}llll@{}}
\toprule
\textbf{Annotation stage} & \textbf{Human annotation} & \textbf{IAA (human--human)} & \textbf{Judge validation} \\
\midrule
Stance labeling & 1 author, 20 essays & -- & vs.\ author: 70.0\% ($\kappa=0.625$) \\
Main-/sub-argument extraction & 1 author, 30 essays & -- & 30 main, 134 sub-arguments \\
Pairwise argument overlap & 2 authors, 100 pairs & $\kappa=0.61$\,/\,$0.80$ & vs.\ humans: 93\% (coarse) \\
Paragraph discourse mode & 2 authors, 83 paragraphs & 84.1\% ($\kappa=0.49$) & vs.\ gold: 91.5\% ($\kappa=0.71$) \\
Paragraph argument role & 2 authors, 83 paragraphs & 54.2\% ($\kappa=0.49$) & vs.\ gold: 51.3\% ($\kappa=0.44$) \\
\bottomrule
\end{tabular*}
\caption{\textbf{Human validation across annotation stages:} annotation effort, human--human agreement (IAA), and the LLM judge's agreement with human labels. Pairwise-overlap $\kappa$ values are exact four-way\,/\,coarse agreement (\autoref{tab:pairwise-validation}); paragraph-level judge results are measured against an adjudicated gold set, with exact match over label sets for the multi-label argument role.}
\label{tab:annotation-validation}
\end{table*}

\subsection{Artifact Use and Intended Use}
\label{app:artifact-use}

We use public debate artifacts from \textit{New York Times} Room for Debate and \textit{Boston Review} forums to study how human and LLM-written responses vary when they address the same contested issue. Essays were published for public reading and discussion, and our analysis treats them as public argumentative writing. The essays are written by named public op-ed contributors and forum responders, so we treat authorship as already public. For position-guided generation, we anonymize author bios before passing them to the LLMs.
We do not redistribute the full original human-written essays. We release only research artifacts needed to audit and reproduce the analysis, such as code, prompts, cohort identifiers, URLs, parsed metadata, derived annotations, pairwise overlap labels, aggregate statistics, and model-generated responses. These materials are intended for research only. They should not be used to republish, replace, or commercially redistribute the original NYT or BR texts. LLM-generated essays were produced through official APIs under the relevant provider terms. We did not perform automated offensive-content filtering, because both source corpora are editorially curated.

\subsection{Summary of Human Validation}
\label{app:annotation-validation}

\autoref{tab:annotation-validation} summarizes the human validation performed at each LLM-annotated stage of the pipeline. For the two paragraph-level layers, two authors independently labeled 83 paragraphs on both layers, disagreements were adjudicated into a gold set, and the judge was evaluated against this gold set. The argument-role layer is multi-label, so exact match requires the full label set of a paragraph to agree, which makes agreement substantially harder than in the single-label tasks.

\section{Methodology Appendix}
\label{app:methodology}

\subsection{Model-Call Hyperparameters}
\label{app:model-hyperparameters}

\autoref{tab:model-hyperparameters} reports the generation and annotation settings used across model calls.

\begin{table*}[t]
\centering
\scriptsize
\renewcommand{\arraystretch}{1.12}
\setlength{\tabcolsep}{4pt}
\begin{tabularx}{\textwidth}{@{}L{0.22\textwidth}L{0.28\textwidth}Y@{}}
\toprule
\textbf{Model / tool} & \textbf{Stage / task} & \textbf{Hyperparameters} \\
\midrule
\multicolumn{3}{@{}l}{\textit{Generation}} \\
GPT-5.5 (OpenAI API) &
Essay generation for default (\texttt{v1a}), self-diversified (\texttt{v15a}), and position-guided (\texttt{v4a}) conditions &
\texttt{temperature=1.0}; \texttt{reasoning.effort=medium}; \texttt{top\_p} not set; no web/search tools; NYT runs used provider output defaults, and Boston Review runs used \texttt{max\_output\_tokens=32000}. \\
GPT-5.5 (OpenAI API) &
Self-diversification reasoning-effort sensitivity check &
Same generation settings as above, except \texttt{reasoning.effort=xhigh}. \\
Gemini-3.1-Pro-Preview (Vertex) &
Essay generation for \texttt{v1a}, \texttt{v15a}, and \texttt{v4a} conditions &
\texttt{temperature=1.0}; \texttt{thinking\_level=MEDIUM}; \texttt{top\_p} not set; no web/search tools; NYT runs used provider output defaults, and Boston Review runs used \texttt{max\_output\_tokens=32000}. \\
Claude Opus 4.7 (OpenRouter) &
Essay generation for \texttt{v1a}, \texttt{v15a}, and \texttt{v4a} conditions &
\texttt{temperature=1.0}; \texttt{reasoning.enabled=true}; \texttt{verbosity=high}; \texttt{top\_p} not set; no web/search tools; NYT runs used provider output defaults, and Boston Review runs used \texttt{max\_tokens=32000}. \\
Minimax M2.7 (OpenRouter) &
Essay generation for \texttt{v1a}, \texttt{v15a}, and \texttt{v4a} conditions &
\texttt{temperature=1.0}; \texttt{reasoning.effort=medium}; \texttt{top\_p} not set; no web/search tools; NYT runs used provider output defaults, and Boston Review runs used \texttt{max\_tokens=32000}. \\
DeepSeek V4 Pro (OpenRouter) &
Essay generation for \texttt{v1a}, \texttt{v15a}, and \texttt{v4a} conditions &
\texttt{temperature=1.0}; \texttt{reasoning.effort=medium}; \texttt{top\_p} not set; no web/search tools; NYT runs used provider output defaults, and Boston Review runs used \texttt{max\_tokens=32000}. \\
\midrule
\multicolumn{3}{@{}l}{\textit{Annotation and preprocessing}} \\
Gemini-3-Flash-Preview (Vertex) &
Topic, question-type, sensitivity, and temporal-change tags for sampling and filtering &
\texttt{temperature=0.0}; \texttt{thinking\_level=MINIMAL}; \texttt{max\_output\_tokens=400}; \texttt{top\_p} not set; strict JSON post-parse. \\
GPT-5.4-Mini (OpenAI API) &
Temporal-change agreement check &
\texttt{temperature=0.0}; \texttt{reasoning.effort=none}; \texttt{max\_output\_tokens=400}; \texttt{top\_p} not set; strict JSON post-parse. \\
Gemini-3-Flash-Preview (Vertex) &
Position-guide extraction for position-guided generation; Toulmin-style extraction of arguments and sub-arguments &
\texttt{temperature=0.0}; \texttt{thinking\_level=MINIMAL}; \texttt{max\_output\_tokens=1200}; \texttt{top\_p} not set; strict JSON post-parse. \\
Gemini-3-Flash-Preview (Vertex) &
Pairwise main-argument overlap judgment &
\texttt{temperature=0.0}; \texttt{thinking\_level=MINIMAL}; \texttt{max\_output\_tokens=2000}; \texttt{top\_p} not set; strict JSON post-parse\\
Gemini-3-Flash-Preview (Vertex) &
Pairwise sub-argument overlap judgment, including the cross-stance reuse analysis &
\texttt{temperature=0.0}; \texttt{thinking\_level=MINIMAL}; \texttt{max\_output\_tokens=1800}; \texttt{top\_p} not set; strict JSON post-parse. \\
Gemini-3-Flash-Preview (Vertex) &
Topic-agnostic bucket-register contrastive descriptions &
\texttt{temperature=0.0}; \texttt{thinking\_level=MINIMAL}; \texttt{top\_p} not set; paragraph-form response. \\
Gemini-3-Flash-Preview (Vertex) &
Stance-axis extraction and five-point essay stance labeling &
\texttt{response\_mime\_type=application/json}; \texttt{thinking\_level=MINIMAL}; stage-1 \texttt{max\_output\_tokens=400}; stage-2 \texttt{max\_output\_tokens=300}; \texttt{top\_p} not set. \\
Gemini-3-Flash-Preview (Vertex) &
Paragraph-level argumentative-role and discourse-mode annotation &
\texttt{response\_mime\_type=application/json}; \texttt{thinking\_level=MINIMAL}; \texttt{max\_output\_tokens=4000}; \texttt{top\_p} not set. \\
\bottomrule
\end{tabularx}
\caption{\textbf{Model settings used for generation and annotation.} Provider defaults mean that a parameter was not explicitly set in the request. No web-search or retrieval tools were enabled. }
\label{tab:model-hyperparameters}
\end{table*}

\subsection{Content Metric Details}
\label{app:content-metrics}

\autoref{tab:content-metrics-details} summarizes the content metrics used in the main-argument and sub-argument analyses. All are computed from the same pairwise overlap labels and use \texttt{equivalent} or \texttt{strong\_overlap} as the substantial-overlap boundary unless otherwise specified.

\begin{table*}[t]
\centering
\small
\renewcommand{\arraystretch}{1.14}
\setlength{\tabcolsep}{5pt}
\begin{tabularx}{\textwidth}{@{}L{0.22\textwidth}Y@{}}
\toprule
\textbf{Metric} & \textbf{Definition} \\
\midrule
Within-group unique rate $U_m$ & Expected fraction of a group's argument units that have \emph{no} substantial-overlap match with another unit from the same group and debate in a same-sized sample of size $m$. This is the primary metric for main-argument and sub-argument collapse. \\
Human-cluster recovery & Share of distinct human main-argument clusters in a debate that are substantially overlapped by at least one LLM-generated main argument. This asks how much of the observed human argument space LLMs reach. \\
Generated-side human overlap & Share of LLM-generated main-argument clusters that substantially overlap at least one human main-argument cluster in the same debate. This asks how much generated variation falls inside the observed human argument space. \\
Cluster-size bands & Descriptive grouping of argument clusters by how many units they contain. We use this in appendix analyses to ask whether common human arguments are easier to recover than one-off human arguments. \\
Cluster-region share $\rho$ & For a sub-argument cluster, $\rho = n_\mathrm{LLM}/(n_\mathrm{LLM}+n_\mathrm{Human})$ is the share of units contributed by LLMs. This supports the appendix analysis of human-dominant, mixed, and LLM-dominant sub-argument regions. \\
Symmetric reuse / recovery $r(A,B)$ & Pair- or pool-level overlap between two essays or writer pools, defined as the average of directional recovery from $A$ to $B$ and from $B$ to $A$. This is used for sub-argument cross-group and cross-stance analyses, with common-size subsampling where pool sizes differ. \\
\bottomrule
\end{tabularx}
\caption{\textbf{Content metrics used in the analysis.} Metrics are derived from pairwise argument-overlap labels. $U_m$ measures within-group uniqueness; recovery and generated-side overlap compare human and LLM argument spaces; cluster-size and $\rho$ summaries support appendix analyses of which arguments are recovered or repeatedly reused; $r(A,B)$ measures pair- or pool-level sub-argument reuse.}
\label{tab:content-metrics-details}
\end{table*}

\subsection{Pairwise Argument-Overlap Validation}
\label{app:pairwise-validation}

\begin{figure*}[!htbp]
\centering
\includegraphics[width=0.95\textwidth,keepaspectratio]{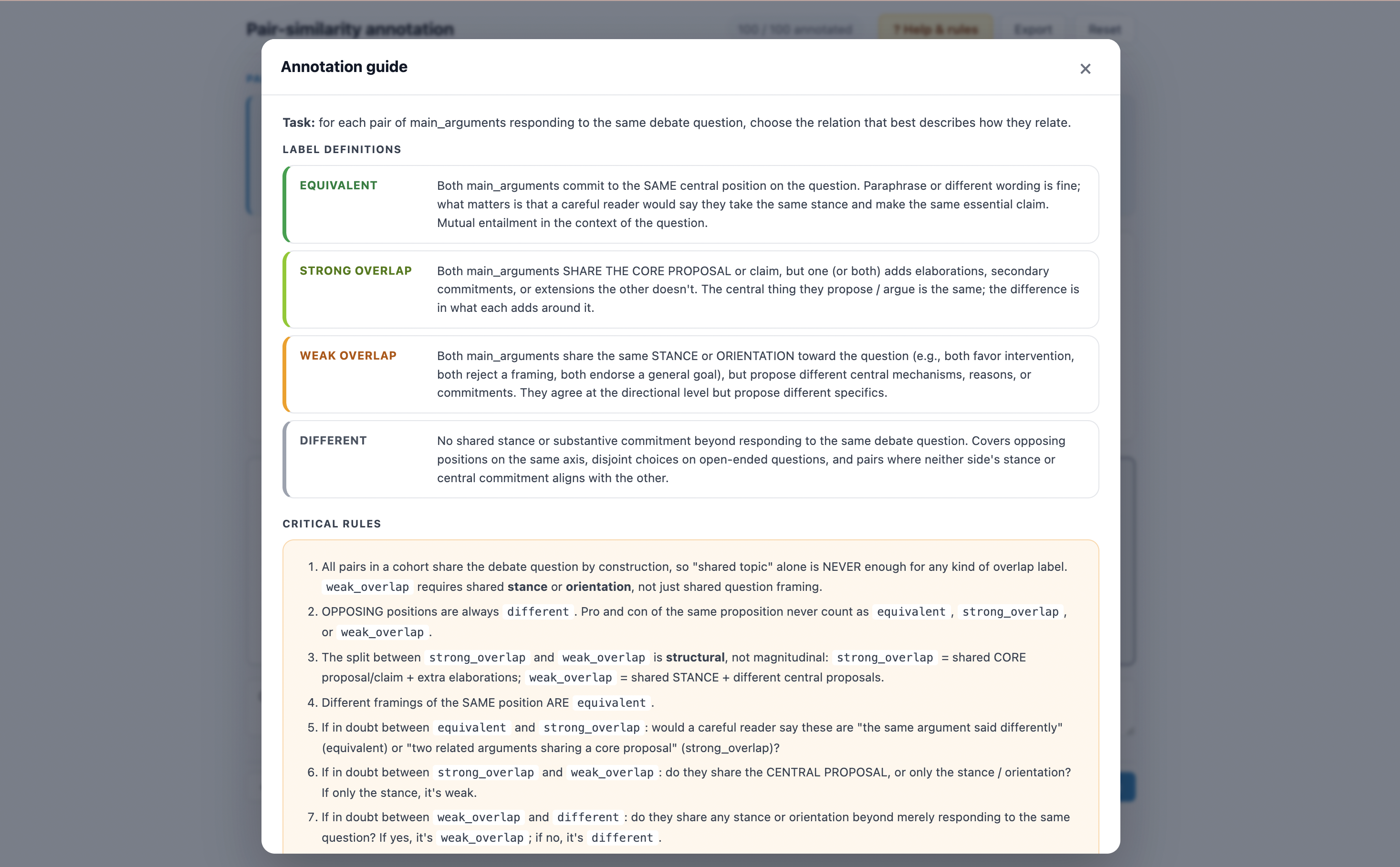}
\\[6pt]
\includegraphics[width=0.95\textwidth,keepaspectratio]{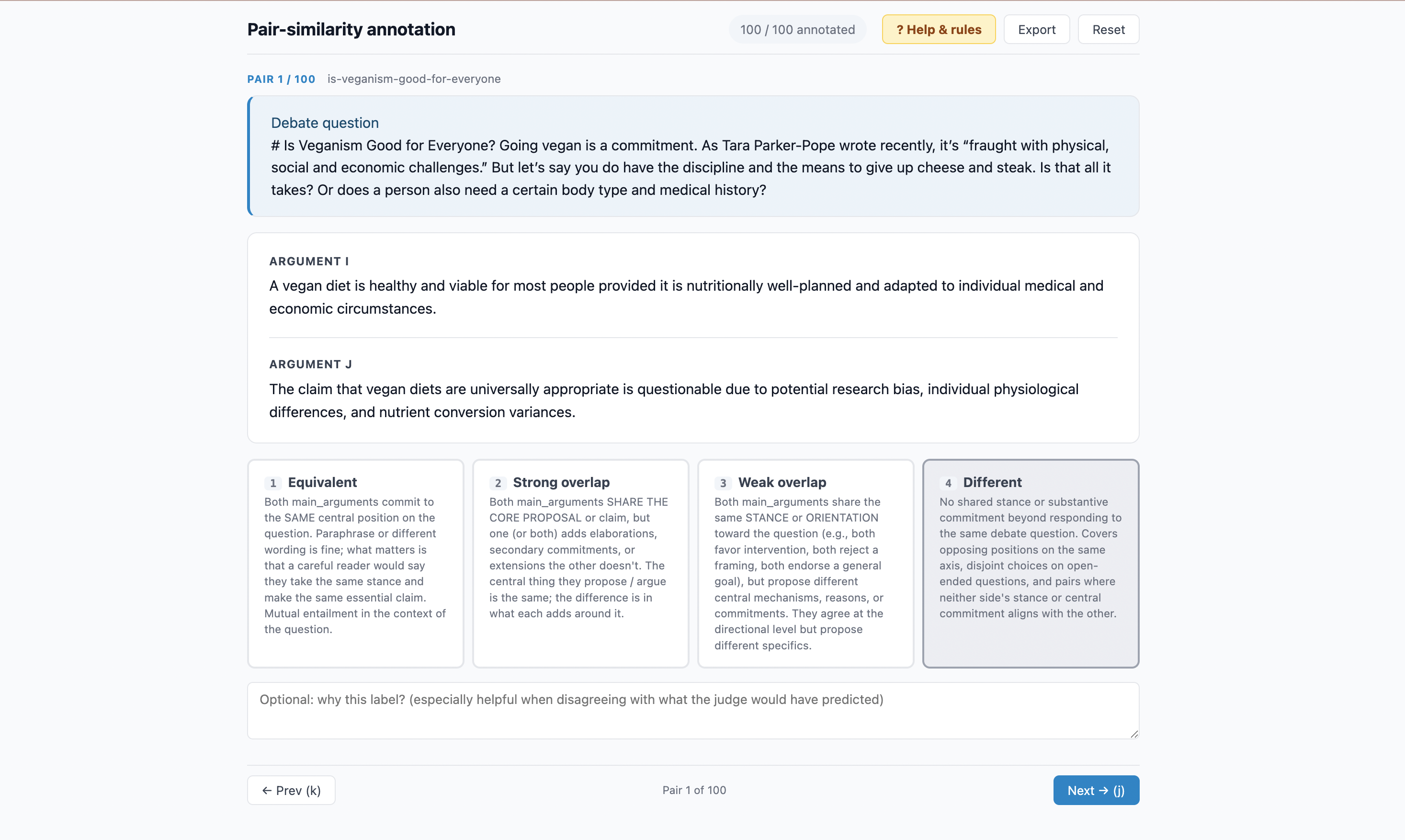}
\caption{\textbf{Annotation rules (top) and interface (bottom) for the pairwise argument-overlap task.} Each annotator received the four-label rubric, definitions, and decision guidance before labeling, and selected one of \texttt{equivalent}, \texttt{strong\_overlap}, \texttt{weak\_overlap}, or \texttt{different} for each pair.}
\label{fig:pair-judge-annotation}
\end{figure*}

We manually validated the four-label pairwise schema used for main-argument overlap on a set of 100 same-debate argument pairs from the analysis data. Two authors independently labeled all 100 pairs using the same four labels used by the automatic judge: \texttt{equivalent}, \texttt{strong\_overlap}, \texttt{weak\_overlap}, and \texttt{different}. Annotators followed the labeling rules and used the interface shown in \autoref{fig:pair-judge-annotation}. The schema was evaluated at two resolutions. The fine evaluation requires an exact match on the four-way label. The coarse evaluation asks whether raters agree on the analysis-critical boundary between substantially overlapping arguments (\texttt{equivalent} or \texttt{strong\_overlap}) and arguments that are non-overlapping or only weakly related (\texttt{weak\_overlap} or \texttt{different}).

\begin{table*}[t]
\centering
\small
\renewcommand{\arraystretch}{1.12}
\setlength{\tabcolsep}{5pt}
\begin{tabular*}{\textwidth}{@{\extracolsep{\fill}}lcc@{}}
\toprule
Comparison & Fine agreement & Coarse agreement \\
\midrule
Judge--Author 1 & 69\% ($\kappa=0.58$) & 93\% ($\kappa=0.86$) \\
Judge--Author 2 & 63\% ($\kappa=0.50$) & 93\% ($\kappa=0.86$) \\
Author 1--Author 2 & 72\% ($\kappa=0.61$) & 90\% ($\kappa=0.80$) \\
\bottomrule
\end{tabular*}
\caption{\textbf{Manual validation of pairwise main-argument labels.} Fine agreement is exact agreement on the four-label schema. Coarse agreement groups \texttt{equivalent}/\texttt{strong\_overlap} as substantially overlapping and \texttt{weak\_overlap}/\texttt{different} as not substantially overlapping.}
\label{tab:pairwise-validation}
\end{table*}

The validation shows that the hardest boundary is the fine-grained distinction among neighboring labels, especially between \texttt{equivalent} and \texttt{strong\_overlap} or between \texttt{weak\_overlap} and \texttt{different}. Agreement is higher on the broader substantial-overlap boundary, so the primary main-argument cluster analyses use \texttt{equivalent} and \texttt{strong\_overlap} as the merge boundary.

\begin{table*}[t]
\centering
\footnotesize
\renewcommand{\arraystretch}{1.14}
\setlength{\tabcolsep}{3.5pt}
\begin{tabularx}{\textwidth}{@{}L{0.13\textwidth}L{0.18\textwidth}YYL{0.16\textwidth}@{}}
\toprule
\textbf{Label} & \textbf{Debate} & \textbf{Argument A} & \textbf{Argument B} & \textbf{Interpretation} \\
\midrule
\texttt{equivalent} &
Privacy and the Internet of Things &
To realize the benefits of the Internet of Things, manufacturers and regulators must prioritize security by design, transparency, and consumer privacy protections. &
To realize the Internet of Things' potential, manufacturers and regulators must implement security-by-design standards and transparent data controls to ensure user trust. &
Same core solution, stated in different wording. \\
\addlinespace[0.35em]
\texttt{strong\_overlap} &
Too Few Fish in the Sea &
Sustainable seafood is achievable only by redefining market demand through transparency, science-based regulation, and shifting consumption to lower-impact species. &
Ethical seafood consumption requires shifting demand toward lower-trophic species and freshwater fish rather than relying on flawed eco-labeling of popular, overfished predatory species. &
Shared core proposal, with one argument adding broader regulation and transparency. \\
\addlinespace[0.35em]
\texttt{weak\_overlap} &
Is Veganism Good for Everyone? &
A well-planned vegan diet is healthy for most people, but its success depends on nutritional knowledge, supplementation, and individual medical circumstances. &
A strictly vegan diet is not universally optimal because individual biological diversity and medical conditions make it practically or biologically unsuitable for everyone. &
Same orientation, but different central claim. \\
\addlinespace[0.35em]
\texttt{different} &
Can the Market Stave Off Global Warming? &
Cap-and-trade systems are an ineffective and inequitable mechanism for addressing climate change, particularly regarding the needs and economic realities of developing nations. &
A national carbon-pricing regime, specifically cap-and-trade, is the only feasible and effective method for the U.S. to achieve significant long-term emission reductions. &
Opposing positions on the debate question. \\
\bottomrule
\end{tabularx}
\caption{\textbf{Examples of pairwise main-argument overlap labels from the validation set.} Rows show actual same-debate main-argument pairs for which the final judge label and human validation label agreed. These examples were not used as calibration examples in the judge prompt.}
\label{tab:pairwise-label-examples}
\end{table*}

During schema development, we also used repeated judge revisions to resolve systematic errors. One revision removed calibration examples drawn from real debates in the analysis set and replaced them with invented examples. Another targeted open-ended debates where two arguments merely shared the debate's broad goal but proposed different central mechanisms; those cases had been over-labeled as \texttt{weak\_overlap}, and the revised schema directed them to \texttt{different}.

\section{Main-Argument Appendix}
\label{app:main-arg}


\begin{table*}[!t]
    \centering
    \small
    \renewcommand{\arraystretch}{1.4}
    \setlength{\tabcolsep}{8pt}
    \begin{tabularx}{\textwidth}{@{}>{\raggedright\arraybackslash}X r@{}}
    \toprule
    \textbf{Example main argument from the ``What's lost and gained as Silicon Valley shapes Washington?'' debate}
     & \makecell[r]{\textbf{\textcolor{divcolor}{\divshort}}\\ {\scriptsize families}} \\
    \midrule
    \rowcolor{secbg}\textbf{\textcolor{secfg}{(1)\,\,\defaultshort collapses onto a hedged frame, also present in all \divshort pools}} & \textbf{\textcolor{secfg}{5/5}}\\
    \textcolor{vancolor}{\footnotesize\textsc{GPT}}\enspace\textit{While Silicon Valley's influence can \uline{modernize government competence and technical literacy}, it must be balanced with rigorous oversight to prevent \textbf{corporate interests from undermining democratic accountability and public interest}.}
     & \\
    \textcolor{vancolor}{\footnotesize\textsc{Claude}}\enspace\textit{While the influx of Silicon Valley expertise significantly \uline{improves the efficiency and usability of government digital services}, it simultaneously risks \textbf{compromising regulatory integrity and democratic deliberation through corporate lobbying} and a bias toward technological disruption.}
     & \\
    \textcolor{vancolor}{\footnotesize\textsc{MiniMax}}\enspace\textit{While the influx of Silicon Valley talent can \uline{improve government operational efficiency}, it poses significant risks to \textbf{democratic legitimacy and the prioritization of public values over corporate interests}.}
     & \\
    \textcolor{vancolor}{\footnotesize\textsc{DeepSeek}}\enspace\textit{While Silicon Valley's influence in Washington provides \uline{essential modernization of public services}, it simultaneously threatens \textbf{democratic accountability by prioritizing technocratic efficiency over regulatory independence and public interest}.}
     & \\
    \textcolor{vancolor}{\footnotesize\textsc{Gemini}}\enspace\textit{While the integration of Silicon Valley talent is \uline{essential for modernizing government services}, the increasing \textbf{political influence and lobbying power of tech giants threaten objective regulatory oversight and democratic protections}.}
     & \\
    \midrule
    \rowcolor{secbg}\multicolumn{2}{@{}l@{}}{\textbf{\textcolor{secfg}{(2)\,\,\divshort partially recovers a human argument}}}\\
    \textit{The increasing influence of Silicon Valley creates a dangerous dependency that \textbf{shifts power from state agencies to private contractors}, undermining the government's ability to regulate the tech industry effectively.}
     & 2/5 \\
    \midrule
    \rowcolor{secbg}\multicolumn{2}{@{}l@{}}{\textbf{\textcolor{secfg}{(3a)\,\,But \divshort misses more distinctive human arguments\dots}}}\\
    \textit{While Silicon Valley offers valuable technical skills, its \textbf{lack of demographic diversity} creates significant blind spots that prevent it from effectively serving the needs of all citizens.}
     & 0/5 \\
    \textit{The expansion of the H-1B visa program, driven by Silicon Valley's lobbying, should be treated with skepticism because the economic benefits are overstated and the program often \textbf{displaces American workers}.}
     & 0/5 \\
    \textit{The influence of Silicon Valley in Washington is \textbf{not fundamentally different from that of other innovative industries}, and the government should focus on building internal expertise rather than fixating on the novelty of `tech' lobbying.}
     & 0/5 \\
    \midrule
    \rowcolor{secbg}\multicolumn{2}{@{}l@{}}{\textbf{\textcolor{secfg}{(3b)\,\,\dots and introduces arguments no human raised}}}\\
    \textit{The migration of tech talent into the federal government is a \textbf{necessary corrective} that improves the functionality and efficiency of essential public services.}
     & 5/5 \\
    \textit{Silicon Valley's growing influence in Washington is a strategic effort to preserve exploitative business models by \textbf{dismantling federal labor protections} and the social safety net.}
     & 1/5 \\
    \textit{The increasing influence of Silicon Valley risks prioritizing tech-savvy users while \textbf{marginalizing vulnerable populations} who rely on human-centric government services.}
     & 1/5 \\
    \bottomrule
        \end{tabularx}
    \caption{\textbf{Main-argument collapse and an invented consensus: what's lost and gained as Silicon Valley shapes Washington?} Sections follow Table~\ref{tab:main-arg-recovery-examples}. \textbf{(1)}~All five vanilla (\defaultshort) outputs cite the same gain (\textbf{modernizing government services}) and the same loss (\textbf{threatening democratic accountability}), with all five opening ``While\dots''. This single hedge is the only argument vanilla produces, and \divshort\ never escapes it either ($5/5$). Compared with the cleanliness case(Table~\ref{tab:main-arg-recovery-examples}), diversification recovers less: only one human argument is partially recovered ($2/5$ in
  \textbf{(2)}), while \textbf{(3a)} shows three distinctive humans entirely missed ($0/5$ each). \textbf{(3b)} shows a second convergence: all five families produce an invention no human raised, ``tech-talent migration is a \textbf{necessary corrective} for failing public
  services'' ($5/5$). Other inventions stay at single-family margins. Diversification reproduces vanilla's convergence, just outside the human distribution.}
    \label{tab:main-arg-recovery-siliconvalley}
    \vspace{-10pt}
\end{table*}

\subsection{Self-Diversification Recovery Details}
\label{app:main-arg-recovery}

\begin{figure}[t]
\centering
\includegraphics[width=\columnwidth]{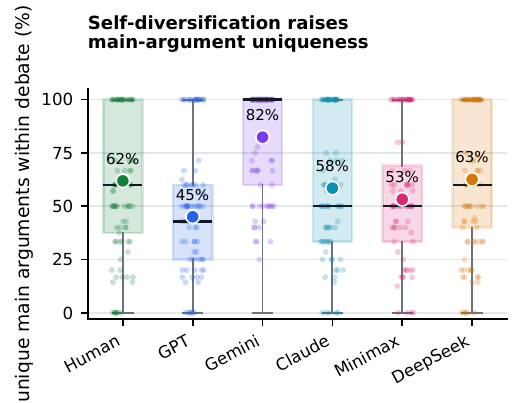}
\caption{\textbf{Diversified prompting raises within-model main-argument uniqueness.} Share of main arguments that are unique within the debate for all human writers and for medium-effort diversified outputs from each LLM family. Small points show debate-level observations; large points show group means.}
\label{fig:main-arg-diversity-recovery}
\end{figure}

\autoref{fig:main-arg-diversity-recovery} visualizes uniqueness within each model's diversified outputs, and \autoref{tab:main-arg-recovery} reports the corresponding means and confidence intervals. This is a within-model comparison: a diversified main argument is counted as unique if it does not substantially overlap another medium-effort output from the same model in the same debate. We use the same substantial-overlap boundary as the default analysis.

\begin{table*}[t]
\centering
\small
\renewcommand{\arraystretch}{1.1}
\setlength{\tabcolsep}{6pt}
\begin{tabular*}{\textwidth}{@{\extracolsep{\fill}}lcc@{}}
\toprule
Group & Unique main arguments & Difference from human mean \\
\midrule
Human writers & $61.9\%$ \scriptsize{[57.5, 66.2]} & --- \\
GPT & $45.0\%$ \scriptsize{[40.9, 49.2]} & $-16.8$ pp \\
Gemini & $82.4\%$ \scriptsize{[79.0, 85.7]} & $+20.5$ pp \\
Claude & $58.5\%$ \scriptsize{[54.0, 63.0]} & $-3.4$ pp \\
Minimax & $53.2\%$ \scriptsize{[48.8, 57.4]} & $-8.7$ pp \\
DeepSeek & $62.6\%$ \scriptsize{[58.0, 67.2]} & $+0.7$ pp \\
\bottomrule
\end{tabular*}
\caption{\textbf{Main-argument uniqueness under diversified prompting.} Values report the average share of main arguments that are unique within the debate, with 95\% cluster-bootstrap CIs in brackets. The human row uses all human writers in each debate, while model rows use medium-effort diversified outputs from one model at a time. This differs from the vanilla comparison in \S\ref{sec:main_arg}, which uses a common-size comparison between human writers and five LLM representatives; here the goal is to measure how much diversity each model produces within its own diversified outputs.}
\label{tab:main-arg-recovery}
\end{table*}

\subsection{Which Human Main Arguments Are Recovered?}
\label{app:main-arg-recovery-types}

We also aggregate recovery at the level of human main-argument clusters. A cluster is counted as recovered if at least one medium-effort self-diversified output from any of the five LLMs substantially overlaps that human cluster (\texttt{equivalent} or \texttt{strong\_overlap}). Under this pooled view, $73.9\%$ of human clusters are recovered by at least one LLM, and the average human cluster is recovered by $2.45$ of the five LLMs.

The strongest pattern is cluster popularity. Human arguments made by multiple writers in the same debate are recovered $98.1\%$ of the time. Human arguments made by a single writer are recovered less often ($67.8\%$). Recovery is also higher for binary debates than open-ended debates ($79.0\%$ versus $69.7\%$). A qualitative inspection of recovered and missed clusters suggests that LLMs most reliably recover broad, direct answers to the debate question, while missed clusters more often involve specific examples, narrower proposals, or author-specific framings. \autoref{tab:main-arg-recovery-examples-full} gives additional debate-level examples using the same categories as the main-text table.

\begin{table*}[t]
\centering
\scriptsize
\newcommand{\apphmcell}[2]{\colorbox{#1}{\strut\hspace{2pt}#2\hspace{2pt}}}
\renewcommand{\arraystretch}{1.13}
\setlength{\tabcolsep}{3.5pt}
\begin{tabularx}{\textwidth}{@{}L{0.17\textwidth}L{0.15\textwidth}Y>{\centering\arraybackslash}p{0.075\textwidth}>{\centering\arraybackslash}p{0.075\textwidth}@{}}
\toprule
\textbf{Debate} & \textbf{Argument type} & \textbf{Example main argument} & \textbf{Human side} & \textbf{LLM side} \\
\midrule
Salt shakers on restaurant tables?
& Broad human argument &
Restaurants should provide salt because diners should be able to adjust seasoning to their own palates. &
\apphmcell{teal!25}{$3/6$} &
\apphmcell{teal!35}{$5/5$} \\
\addlinespace[0.25em]
&
Specific human argument &
Salt shakers should be removed because seasoning is the chef's prerogative and restaurants should serve food as the chef intended. &
\apphmcell{teal!18}{$2/6$} &
\apphmcell{teal!35}{$5/5$} \\
\addlinespace[0.25em]
&
LLM-only health argument &
Restaurants should remove salt shakers from tables to improve public health by reducing sodium consumption. &
\apphmcell{gray!10}{$0/6$} &
\apphmcell{teal!25}{$3/5$} \\
\midrule
Is veganism good for everyone?
& Broad human argument &
Veganism may not be suitable for everyone because of individual health conditions, nutritional deficiencies, and limited long-term evidence. &
\apphmcell{teal!35}{$4/6$} &
\apphmcell{teal!35}{$5/5$} \\
\addlinespace[0.25em]
&
Specific human proposal &
Reducing meat consumption is necessary because of the declining quality and safety of industrial animal products, even if total veganism is not for everyone. &
\apphmcell{teal!10}{$1/6$} &
\apphmcell{gray!10}{$0/5$} \\
\addlinespace[0.25em]
&
LLM-only proposal &
Veganism cannot be good for everyone until social and economic infrastructure makes plant-based eating accessible and sustainable for all. &
\apphmcell{gray!10}{$0/6$} &
\apphmcell{teal!35}{$5/5$} \\
\midrule
Should federal money rebuild coastal properties?
& Broad human argument &
Federal disaster spending should shift away from subsidizing reconstruction in high-risk coastal areas and toward relocation and mitigation. &
\apphmcell{teal!35}{$3/5$} &
\apphmcell{teal!35}{$5/5$} \\
\addlinespace[0.25em]
&
Specific human proposal &
The federal government should reduce subsidies and alter land-use policies to limit development in high-risk coastal areas. &
\apphmcell{teal!10}{$1/5$} &
\apphmcell{gray!10}{$0/5$} \\
\addlinespace[0.25em]
&
LLM-only proposal &
Federal rebuilding subsidies should be phased out so private owners bear location risks while public funds are reserved for emergency response and public infrastructure. &
\apphmcell{gray!10}{$0/5$} &
\apphmcell{teal!35}{$5/5$} \\
\midrule
Should overcrowded national parks restrict access?
& Broad human argument &
National parks should prioritize mitigation strategies and funding over broad access restrictions, protecting resources while maintaining public access. &
\apphmcell{teal!10}{$1/4$} &
\apphmcell{teal!35}{$5/5$} \\
\addlinespace[0.25em]
&
Specific human proposal &
The National Park Service should raise entrance fees to reduce overcrowding and fund deferred maintenance. &
\apphmcell{teal!10}{$1/4$} &
\apphmcell{gray!10}{$0/5$} \\
\addlinespace[0.25em]
&
LLM-only proposal &
The National Park Service should use restricted access and reservation systems to prioritize preservation over unrestricted tourist enjoyment. &
\apphmcell{gray!10}{$0/4$} &
\apphmcell{teal!35}{$5/5$} \\
\midrule
Should corporate-funded research be reduced?
& Broad human argument &
Corporate funding of scientific research need not be reduced because integrity depends on research design, execution, and transparency rather than funding source. &
\apphmcell{teal!25}{$2/4$} &
\apphmcell{teal!35}{$5/5$} \\
\addlinespace[0.25em]
&
Specific human proposal &
Corporate funding should not substitute for federal support, because public funding is essential for basic, high-risk research that industry depends on. &
\apphmcell{teal!25}{$2/4$} &
\apphmcell{gray!10}{$0/5$} \\
\addlinespace[0.25em]
&
LLM-only proposal &
Corporate funding of research on public health, safety, and the environment should be sharply reduced to prevent profit-driven distortion of science. &
\apphmcell{gray!10}{$0/4$} &
\apphmcell{teal!35}{$5/5$} \\
\midrule
What is the appeal of astrology?
& Broad human argument &
Astrology appeals not because it is scientifically accurate, but because it offers emotional validation and a framework for self-reflection. &
\apphmcell{teal!10}{$1/5$} &
\apphmcell{teal!35}{$5/5$} \\
\addlinespace[0.25em]
&
Specific human proposal &
Astrology's appeal comes from the human brain's tendency to seek patterns and connections, even when those connections are not empirically valid. &
\apphmcell{teal!25}{$2/5$} &
\apphmcell{gray!10}{$0/5$} \\
\addlinespace[0.25em]
&
LLM-only proposal &
Astrology now functions as a shared language and aesthetic system for social connection and self-expression rather than as a source of prediction. &
\apphmcell{gray!10}{$0/5$} &
\apphmcell{teal!25}{$3/5$} \\
\bottomrule
\end{tabularx}
\caption{\textbf{Additional examples of recovered, missed, and LLM-only main arguments.} Each debate repeats the same three categories used in the main-text table. Counts indicate how many humans or LLMs, respectively, produced a substantially overlapping main-argument cluster.}
\label{tab:main-arg-recovery-examples-full}
\end{table*}

\subsection{Post-Cutoff Boston Review Forums}
\label{app:memorization}

A natural concern is that the debates and forums in our corpora may appear in the models' training data, so the observed convergence could partly reflect memorization of the original discussions rather than generation behavior. Since NYT Room for Debate ended in 2018, we probe this using the six most recent Boston Review forums, published between May 2025 and February 2026. This window is fully post-cutoff for Gemini-3.1-Pro-Preview (training cutoff January 2025), includes one post-cutoff forum and one forum at the cutoff boundary for GPT-5.5 (December 2025), and one post-cutoff forum for Claude Opus 4.7 (January 2026); DeepSeek V4 Pro and Minimax M2.7 do not disclose official cutoffs. All essays are generated without web-search or retrieval tools (\autoref{tab:model-hyperparameters}), so models cannot retrieve the source discussions at inference time.
On these six forums, the collapse pattern is unchanged: $84.0\%$ of human main arguments are unique within a forum, compared with $35.6\%$ for \default LLM arguments, and \diversified prompting recovers only $53.8\%$ of human main-argument clusters, consistent with the full-corpus results (\S\ref{sec:main_arg}). While this check cannot rule out memorization for older debates, it indicates that the collapse we measure is not driven by exposure to the source discussions.

\section{Sub-Argument Appendix}
\label{app:sub-arg}

\subsection{Sub-argument extraction and validation}
\label{app:subarg-extraction}

\paragraph{Sub-argument sampling and validation.}
We analyze a stratified 30-debate subset of the 195-debate sample, balanced across the main-argument convergence spectrum (10 most divergent debates, 10 most convergent, and 10 middle) and covering 8 of 10 topic categories. We focus on this subset because sub-argument analysis requires exhaustive pairwise comparison across extracted supporting claims, producing $294{,}765$ judged pairs at the chosen granularity. At approximately $\sim\$300$ in annotation cost, this set the practical limit for manual evaluation.

Granularity is fixed at the sub-argument level, defined as single supporting claims grounded in 2--5 sentence spans. Finer segmentation would substantially increase the $O(N^2)$ pair count without commensurate analytical benefit, while also removing contextual information needed for meaningful argumentative comparison.

To validate extraction quality, we spot-checked 50 randomly sampled sub-argument extractions against their source essays. Of these, 47 mapped cleanly to source spans and the remaining 3 were lightly paraphrased restatements rather than unsupported abstractions. Given the large within-debate effect sizes observed in the 30-debate subset, scaling the same procedure to the full 195 debates would be expected primarily to narrow confidence intervals rather than change the qualitative direction of the results.

\begin{figure*}[t]
\centering
\includegraphics[width=\textwidth]{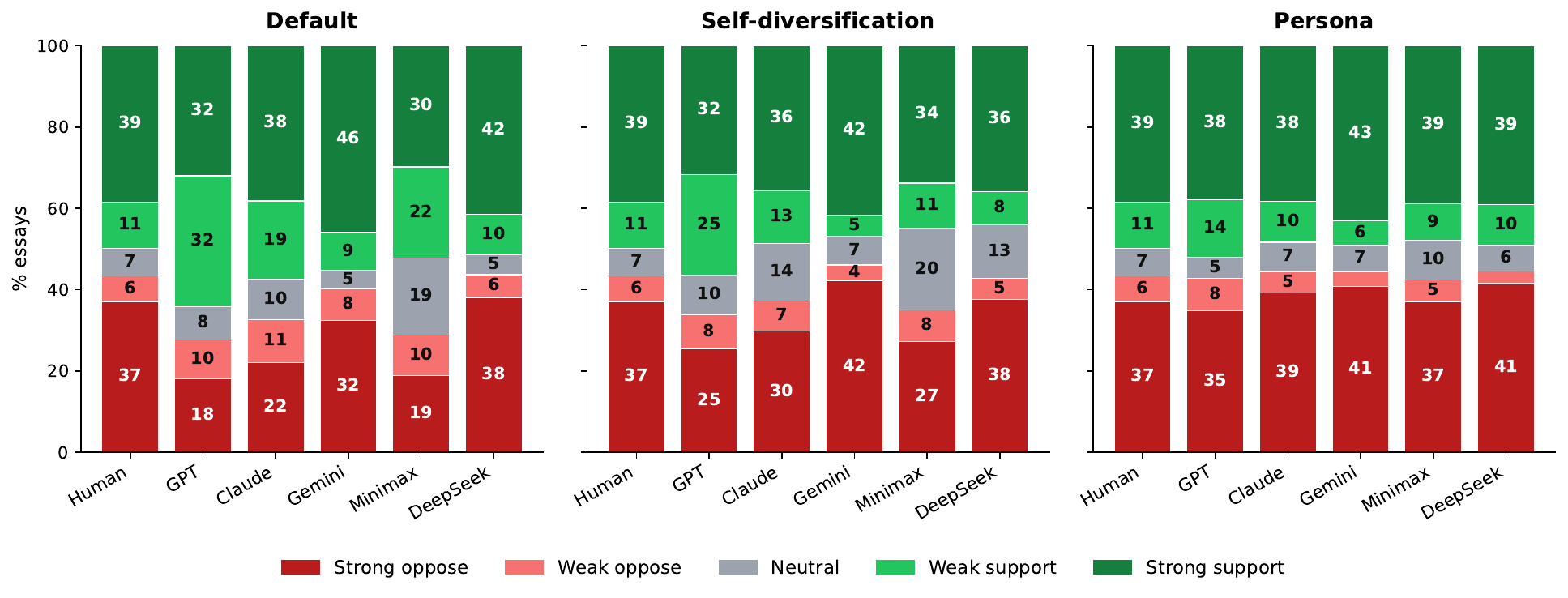}
\caption{\textbf{Stance distribution per prompt, by family.} Five-point stance label distribution for the binary cohorts in the analysis sample ($n=8{,}496$ essays). Each panel is one LLM prompt condition (default, self-diversified, position-guided); within each panel, the leftmost bar is the human reference and the remaining bars are the five LLM families.}
\label{fig:stance-by-prompt-family}
\end{figure*}

\paragraph{Stance distribution per prompt and family.}
\S\ref{fig:stance-by-prompt-family} reports the full five-point stance label distribution for the 97 binary cohorts in the analysis sample, broken down by prompt condition and model family.

\paragraph{Pairwise label transitivity.}
The four-label pair judge produces graded similarity scores rather than a formal equivalence relation, so we report the empirical transitivity of its labels on the $16$-cohort subset. \autoref{tab:app-transitivity} reports the label distribution on the third edge $(A,C)$ given that both anchor edges $(A,B)$ and $(B,C)$ fall above a chosen threshold. Equivalence is not strictly transitive ($67.3\%$ at the strict anchor), but the labels behave as a well-ordered graded similarity scale: chained near-matches stay near and \emph{never} fall to \texttt{unrelated}. This is also why our sub-argument clusters are built as connected components over \texttt{equivalent} (strict) or $\{\texttt{equivalent}, \texttt{strong\_overlap}\}$ (loose) edges rather than as transitive-closure partitions.

\begin{table*}[t]
\centering\small
\renewcommand{\arraystretch}{1.2}
\setlength{\tabcolsep}{5pt}
\begin{tabular}{lcccc r}
\toprule
& \multicolumn{4}{c}{\textbf{Label of $(A,C)$}} & \\
\cmidrule(lr){2-5}
\textbf{Anchor} & \texttt{equiv} & \texttt{strong} & \texttt{weak} & \texttt{unrelated} & \textbf{$n$} \\
\midrule
Strict ($\{\texttt{equiv}\}$)            & $67.3\%$ & $28.2\%$ & $\phantom{0}3.4\%$ & $0\%$ & $19{,}005$ \\
Loose ($\{\texttt{equiv},\texttt{strong}\}$) & $11.0\%$ & $53.0\%$ & $29.3\%$ & $0\%$ & $493{,}602$ \\
\bottomrule
\end{tabular}
\caption{\textbf{Empirical transitivity of pairwise labels.} For each triple $(A,B,C)$ where both anchor edges $(A,B)$ and $(B,C)$ fall above the threshold, we report the label distribution on the third edge $(A,C)$. Annotated triples only (excludes within-essay pairs and missing pairs). $n$ is the number of annotated triples. Labels are short for \texttt{equivalent}, \texttt{strong\_overlap}, \texttt{weak\_overlap}, \texttt{unrelated}.}
\label{tab:app-transitivity}
\end{table*}

\subsection{Within-group unique rate: shared-main-argument subset}
\label{app:shared-main-arg-sampling}

We isolate sub-argument collapse from main-argument variation by restricting to debate questions where humans and default-prompted LLMs independently converge on the same main argument.

\textbf{Cohort selection.} From the $195$-debate sample, we select cohorts that simultaneously satisfy four conditions on the main-argument pairwise judgments (\autoref{sec:content}). Throughout, ``loose'' edges denote \texttt{equivalent} or \texttt{strong\_overlap}, the same boundary used elsewhere in this paper for substantial main-argument overlap.

\begin{enumerate}
\item \emph{V cluster.} For each of the five \default LLM families, we pick a canonical medoid \default essay (the most central essay in that family's modal \texttt{equivalent}-cluster). The five medoids must be transitively connected through loose edges, i.e., the five \default LLMs collectively produce the same main argument.
\item \emph{H cluster.} Among humans, we take the largest connected component under loose human--human edges only (humans must agree among themselves, not through LLM essays). This component must contain at least three humans.
\item \emph{Per-human bridge.} Every human in the H cluster must have at least two loose edges to distinct \default medoids in the V cluster. This per-human requirement rules out cohorts where humans are matched to the LLM cluster only through a single weak bridge.
\item \emph{Diversified coverage.} For each of the five LLM families, at least one \diversified essay must have a loose edge to some human in the H cluster. Enforced at cohort-selection time so that the diversified analysis operates on the same cohort set as the humans/default/position-guided analyses.
\end{enumerate}

The resulting subset contains $16$ debate questions, $62$ cluster humans, and $5 \times 16 = 80$ \default medoids (one per LLM family per cohort). The matching \diversified pool---\diversified essays loose-connected to some human in the H cluster---contains $321$ essays. \autoref{tab:app-shared-main-arg-percohort} reports per-cohort essay counts and average sub-arguments per essay across all writer conditions; \autoref{tab:app-corpus-subarg-counts} reports per-essay sub-argument averages across NYT and BR.

\textbf{Sub-argument labeling.} For each cluster essay we extract sub-arguments and label every inter-essay sub-argument pair on the four-label scale (\autoref{sec:content}). The judged pool covers humans + \default medoids + all \diversified essays matched to the H cluster. Total judged pairs across the $16$ cohorts: $88{,}881$ (humans--humans: $2{,}070$; humans--\default: $3{,}811$; \default--\default: $3{,}074$; \default--\diversified: $18{,}488$; cross-family \diversified--\diversified: $53{,}813$). Same-family \diversified--\diversified pairs are not judged because they never co-occur in a 1-per-family combination.

\textbf{Unique rate $U_m$ and common-$m$ subsampling.} For a comparison pool $P$ in a cohort, $U_m(P)$ is the expected fraction of $P$'s sub-arguments that have \emph{no} same-pool match in a different essay of the same pool (\texttt{equivalent} for strict; \texttt{equivalent} or \texttt{strong\_overlap} for loose). $U_m$ is computed as a closed-form expectation,
\[
  U_m(P) \;=\; \frac{1}{|P|}\sum_{i \in P}
  \frac{\binom{|P|-1-d_i}{\,m-1\,}}{\binom{|P|-1}{\,m-1\,}},
\]
where $d_i$ is the number of within-pool matches argument $i$ has under the chosen threshold. This is the exact expected value under uniform random sampling of $m$ units from $P$, with no Monte Carlo noise. Because pools differ in size, we use a single common-$m$ per cohort to make groups directly comparable.

\textbf{Per-cohort common-$m$.} For each cohort we set
\[
  m_{\mathrm{global}} = \min\bigl(|H|,\; |V|,\; m_{D}^{\min},\; m_{P}^{\min}\bigr),
\]
where $|H|$ and $|V|$ are the sub-argument counts of the H cluster and the \default medoid pool, $m_{P}^{\min}$ is the smallest \persona family pool (under the family-fixed Setup-2 configuration; see below), and $m_{D}^{\min}$ is the smallest \diversified 1-per-family combination pool (sum of the smallest matching \diversified essay per family). Choosing the global minimum guarantees $m$ is feasible for every pool used in the analysis. All conditions are computed at this single $m_{\mathrm{global}}$ per cohort, and per-cohort $U_m$ values are macro-averaged across the $16$ qualifying cohorts; final values appear in \S\ref{tab:app-unique-rates}.

\textbf{Diversified pool construction.} \diversified is the only condition where the relevant pool is itself an average over sampled essay subsets. For each cohort we enumerate all combinations of one \diversified essay per LLM family (each family must contribute at least one essay matched to the H cluster, by condition~(4) above). For each combination, the pool is the union of the five chosen essays' sub-arguments, and we compute $U_{m_{\mathrm{global}}}$ on that pool. The cohort-level \diversified $U_m$ is the unweighted mean of $U_{m_{\mathrm{global}}}$ over all such combinations. Because the closed-form $U_m$ is deterministic, this enumeration produces an exact expected value over the 1-per-family sampling distribution---no Monte Carlo estimation is required.

\textbf{Position-guided matched pools (used in recovery analyses).} For the cross-group recovery analyses (later in this subsection) we extend the shared-main-argument subset to include, for each of the $62$ cluster humans, $5$ \default essays (one per LLM family) and $5$ \persona essays (one per LLM family, guided by that human writer), yielding $310$ \default and $310$ \persona essays. Pool counts in \S\ref{tab:app-shared-main-arg-pools}.

\begin{table}[ht]
\centering\small
\renewcommand{\arraystretch}{1.15}
\begin{tabular}{lrr}
\toprule
\textbf{group}              & \textbf{essays} & \textbf{sub-args} \\
\midrule
human (cluster)             & 62  & 283  \\
\default (position-guided matched)         & 310 & 1{,}367 \\
\persona (position-guided matched)         & 310 & 1{,}388 \\
\midrule
total                       & 682 & 3{,}038 \\
\bottomrule
\end{tabular}
\caption{\textbf{Pool sizes for the shared-main-argument position-guided experiment.} Aggregated across the $16$ shared-main-argument cohorts. Each cluster human contributes one essay per LLM condition per family ($5$ families); \persona essays are generated from that human writer's anonymized position guide.}
\label{tab:app-shared-main-arg-pools}
\end{table}

\begin{table*}[t]
\centering\small
\renewcommand{\arraystretch}{1.1}
\setlength{\tabcolsep}{4pt}
\begin{tabular}{lrrrrrrrr}
\toprule
& \multicolumn{2}{c}{\textbf{Humans}} & \multicolumn{2}{c}{\textbf{\default}} & \multicolumn{2}{c}{\textbf{\diversified}} & \multicolumn{2}{c}{\textbf{\persona}} \\
\cmidrule(lr){2-3}\cmidrule(lr){4-5}\cmidrule(lr){6-7}\cmidrule(lr){8-9}
\textbf{Cohort} & $n$ & avg & $n$ & avg & $n$ & avg & $n$ & avg \\
\midrule
inspector-general-police                & 4 & 5.00 & 5 & 5.00 & 26 & 4.77 & 20 & 4.65 \\
social-networks-fad                     & 7 & 4.71 & 5 & 4.40 & 25 & 4.16 & 35 & 4.31 \\
cyclists-drivers-rules                  & 3 & 4.00 & 5 & 4.20 & 14 & 4.71 & 15 & 4.80 \\
government-wildfires                    & 3 & 4.33 & 5 & 4.40 & 24 & 4.42 & 15 & 4.40 \\
hiring-surrogacy                        & 3 & 5.00 & 5 & 4.20 & 13 & 4.77 & 15 & 4.87 \\
young-people-vote                       & 6 & 3.83 & 5 & 4.00 & 18 & 4.06 & 30 & 3.87 \\
postpartum-depression                   & 3 & 4.33 & 5 & 4.20 & 11 & 4.64 & 15 & 4.53 \\
natural-disasters-acts-god              & 3 & 4.33 & 5 & 4.00 & 19 & 4.32 & 15 & 4.27 \\
robert-durst-forensics                  & 3 & 5.00 & 5 & 5.20 & 17 & 4.76 & 15 & 4.73 \\
drug-enforcement-states                 & 3 & 4.33 & 5 & 4.40 & 24 & 4.79 & 15 & 4.40 \\
coastal-properties                      & 4 & 5.00 & 5 & 4.40 & 14 & 4.50 & 20 & 4.60 \\
cybersecurity-mandates                  & 4 & 4.75 & 5 & 5.00 & 25 & 4.92 & 20 & 5.15 \\
basketball-rim                          & 5 & 4.00 & 5 & 3.60 & 25 & 4.28 & 25 & 3.84 \\
purebred-dogs                           & 3 & 4.33 & 5 & 4.20 & 12 & 4.50 & 15 & 4.33 \\
sex-offenders-restrictions              & 4 & 5.25 & 5 & 4.40 & 26 & 4.96 & 20 & 4.90 \\
safer-if-fewer-jailed                   & 4 & 5.00 & 5 & 4.40 & 28 & 4.93 & 20 & 4.70 \\
\midrule
\textbf{TOTAL / avg}                    & \textbf{62} & 4.58 & \textbf{80} & 4.38 & \textbf{321} & 4.59 & \textbf{310} & 4.52 \\
\bottomrule
\end{tabular}
\caption{\textbf{Per-cohort essay counts and average sub-arguments per essay across the $16$ shared-main-argument cohorts.} For each writer condition, $n$ is the number of essays in that cohort and avg is the mean sub-argument count per essay. \default uses one canonical medoid per LLM family (5 per cohort). \diversified counts include all \diversified essays loose-matching some human in the H cluster. \persona is $5\times$ the cluster human count (one \persona essay per LLM family per cluster human). Average sub-argument counts are tightly comparable across conditions (per-cohort range $3.6$--$5.3$).}
\label{tab:app-shared-main-arg-percohort}
\end{table*}

\begin{table*}[t]
\centering\small
\renewcommand{\arraystretch}{1.20}
\setlength{\tabcolsep}{8pt}
\begin{tabular}{lcc}
\toprule
\textbf{Group} & \textbf{Strict (\texttt{equivalent})} & \textbf{Loose (\texttt{equivalent} or \texttt{strong\_overlap})} \\
\midrule
Humans (cluster)                                    & $94.9\%$ & $41.0\%$ \\
\default LLMs                & $60.6\%$ & $\phantom{0}9.1\%$ \\
\diversified LLMs (1-per-family, Method 2)          & $81.0\%$ & $22.9\%$ \\
Same human writer, different LLMs        & $56.4\%$ & $\phantom{0}6.8\%$ \\
Different human writers, same LLM         & $72.7\%$ & $18.4\%$ \\
\bottomrule
\end{tabular}
\caption{\textbf{Within-group sub-argument unique rates across conditions.} Per-cohort macro-averaged $U_m$ under strict (\texttt{equivalent} only) and loose (\texttt{equivalent} or \texttt{strong\_overlap}) thresholds. All values are common-$m$ subsampled at the cohort level. Per-family breakdowns appear in \autoref{tab:app-per-family-unique}.}
\label{tab:app-unique-rates}
\end{table*}

\begin{table*}[t]
\centering\small
\renewcommand{\arraystretch}{1.15}
\setlength{\tabcolsep}{8pt}
\begin{tabular}{lcccc}
\toprule
 & \multicolumn{2}{c}{\textbf{Within-LLM (family-fixed)}} & \multicolumn{2}{c}{\textbf{Default (vs.\ other families)}} \\
\cmidrule(lr){2-3}\cmidrule(lr){4-5}
\textbf{LLM family} & Strict & Loose & Strict & Loose \\
\midrule
GPT      & $69.4\%$ & $15.3\%$ & $50.2\%$ & $\phantom{0}6.2\%$ \\
Claude   & $81.3\%$ & $21.4\%$ & $50.6\%$ & $\phantom{0}7.0\%$ \\
Gemini   & $64.4\%$ & $16.7\%$ & $38.1\%$ & $\phantom{0}1.6\%$ \\
Minimax  & $77.3\%$ & $22.2\%$ & $63.7\%$ & $\phantom{0}9.1\%$ \\
DeepSeek & $71.3\%$ & $16.3\%$ & $44.8\%$ & $\phantom{0}2.8\%$ \\
\bottomrule
\end{tabular}
\caption{\textbf{Per-family unique rates.} \emph{Within-LLM (family-fixed)}: $U_m$ across position-guided essays for different human writers within each LLM family. \emph{Default (vs.\ other families)}: for each family $f$ and cohort, fraction of $f$'s \default sub-arguments with no equivalent (strict) or equivalent/strong-overlap (loose) match in the union of the other four families' \default sub-arguments, macro-averaged across the $16$ cohorts. Low Default values indicate that the family's \default reasoning is largely reused by other families' \default reasoning.}
\label{tab:app-per-family-unique}
\end{table*}

\paragraph{Cross-corpus replication: Boston Review modal-main-argument subset.}
\label{app:BR-shared-main-arg-sampling}
We apply the within-group $U_m$ analysis to the $61$-forum Boston Review corpus described in \autoref{sec:data}. The NYT shared-main-argument filter from \S\ref{app:shared-main-arg-sampling} (which requires humans and all five \default families to converge on the \emph{same} main-argument cluster) is too strict for BR: BR responses develop more fine-grained main arguments and each forum contains fewer essays per group, so cross-group convergence on a single main-argument cluster is rare. We therefore use a per-group modal-cluster variant. For each forum we independently identify (i)~the largest connected component of humans under loose human--human main-argument overlap, and (ii)~the largest connected component of \default essays under loose \default--\default overlap. From the \default cluster we select up to five canonical medoids using the same \texttt{select\_llm\_representatives} procedure as for NYT. Forums qualify if both the human cluster and the resulting medoid set contain at least three essays; this yields $16$ qualifying forums, $60$ cluster human responses, and $70$ \default medoid responses (avg.\ $3.8$ humans + $4.4$ medoids per forum). Note that the human and \default modal clusters in a given forum may correspond to different main arguments---this is a deliberate consequence of the per-group choice, motivated by BR's finer-grained main-argument distribution. We label every inter-essay sub-argument pair on the four-label scale and compute within-group unique rate $U_m$ per forum under the same common-$m$ procedure as NYT ($m = \min(|H|, |V|)$), then macro-average across forums. Loose-threshold rates: $U_\text{human} = 42.2\%$ and $U_\text{vanilla} = 16.3\%$ (NYT $41.0\%$ and $9.1\%$). Strict-threshold rates: $U_\text{human} = 89.8\%$, $U_\text{vanilla} = 56.8\%$ (NYT $94.9\%$ and $60.6\%$). The loose \default rate is higher on BR than NYT ($16.3\%$ vs.\ $9.1\%$), consistent with BR responses' finer-grained main-argument distribution and smaller per-cluster pool sizes producing more within-cluster heterogeneity. The LLM--human gap is preserved in both loose ($-25.9$pp BR vs.\ $-31.9$pp NYT) and strict thresholds ($-33.0$pp BR vs.\ $-34.3$pp NYT), so sub-argument collapse is robust across corpora with very different essay lengths and forum styles. Per-forum BR breakdowns appear in \autoref{tab:app-BR-percohort}.

\begin{table}[ht]
\centering\small
\renewcommand{\arraystretch}{1.15}
\setlength{\tabcolsep}{6pt}
\begin{tabular}{llrrr}
\toprule
\textbf{Corpus} & \textbf{Group} & \textbf{Essays} & \textbf{Sub-args} & \textbf{Avg / essay} \\
\midrule
NYT (16 cohorts) & humans       & $62$  & $283$ & $4.56$ \\
NYT (16 cohorts) & \default     & $80$  & $350$ & $4.38$ \\
\addlinespace[1pt]
BR (16 forums)   & humans       & $60$  & $311$ & $5.18$ \\
BR (16 forums)   & \default     & $70$  & $348$ & $4.97$ \\
\bottomrule
\end{tabular}
\caption{\textbf{Average sub-argument count per essay across corpora.} Humans and \default produce comparable numbers of sub-arguments per essay within each corpus; BR essays carry slightly more sub-arguments on average, consistent with their longer response length.}
\label{tab:app-corpus-subarg-counts}
\end{table}

\begin{table*}[t]
\centering\small
\renewcommand{\arraystretch}{1.1}
\setlength{\tabcolsep}{4pt}
\begin{tabular}{lrrrrrr}
\toprule
& \multicolumn{3}{c}{\textbf{Humans}} & \multicolumn{3}{c}{\textbf{\default}} \\
\cmidrule(lr){2-4}\cmidrule(lr){5-7}
\textbf{Forum} & $n$ & sub-args & avg & $n$ & sub-args & avg \\
\midrule
after\_neoliberalism          & 4 & 21 & 5.25 & 5 & 26 & 5.20 \\
authentic\_other              & 7 & 35 & 5.00 & 5 & 26 & 5.20 \\
authoritarianism              & 5 & 28 & 5.60 & 4 & 21 & 5.25 \\
campus\_protest               & 3 & 17 & 5.67 & 5 & 24 & 4.80 \\
citizenship\_emergency        & 4 & 22 & 5.50 & 5 & 24 & 4.80 \\
constitutional\_again         & 3 & 16 & 5.33 & 5 & 26 & 5.20 \\
educating\_democracy          & 3 & 16 & 5.33 & 5 & 28 & 5.60 \\
effective\_altruism           & 3 & 14 & 4.67 & 5 & 22 & 4.40 \\
emre\_reproduction            & 3 & 16 & 5.33 & 5 & 25 & 5.00 \\
joseph\_carens                & 3 & 15 & 5.00 & 3 & 15 & 5.00 \\
mlk\_now                      & 3 & 15 & 5.00 & 3 & 14 & 4.67 \\
national\_interest            & 4 & 19 & 4.75 & 4 & 20 & 5.00 \\
neurodiversity                & 3 & 15 & 5.00 & 3 & 15 & 5.00 \\
occupy\_future                & 4 & 19 & 4.75 & 5 & 22 & 4.40 \\
patriotism\_cosmopolitanism   & 4 & 21 & 5.25 & 5 & 26 & 5.20 \\
we\_deserve                   & 4 & 22 & 5.50 & 3 & 14 & 4.67 \\
\midrule
\textbf{TOTAL / avg}          & \textbf{60} & \textbf{311} & 5.18 & \textbf{70} & \textbf{348} & 4.97 \\
\bottomrule
\end{tabular}
\caption{\textbf{Per-forum essay and sub-argument counts in the $16$ BR modal-main-argument forums.} Humans column lists the size of each forum's largest human main-argument cluster; \default lists the canonical medoids selected from the largest \default cluster (up to 5, one per LLM family).}
\label{tab:app-BR-percohort}
\end{table*}

\paragraph{Cross-group recovery analysis.}
\label{app:cross-group-recovery}
Within-group unique rate $U_m$  measures whether a group repeats its own sub-arguments. Here we complement that with \emph{cross-group recovery}: how much of one group's reasoning is reachable from another group's pool. We use the same $16$ shared-main-argument cohorts.

\textbf{Metric.} For pools $A$ and $B$, the recovery rate $R(A \to B)$ is the fraction of $A$'s sub-arguments that share at least one \texttt{equivalent} or \texttt{strong\_overlap} match with some sub-argument in $B$. The symmetric recovery is $r(A,B) = \tfrac{1}{2}(R(A \to B) + R(B \to A))$, with common-$m$ subsampling to control for pool-size asymmetry. We report two views.

\textbf{(i) Pool-to-pool overlap.} \S\ref{tab:app-cross-group-pool} reports $r$ between the cluster-human pool and each LLM-condition pool, macro-averaged across the $16$ cohorts. \persona guidance lifts H$\leftrightarrow$LLM recovery slightly at the strict threshold ($12.8\%$ vs.\ $9.7\%$) but not at the loose threshold ($42.9\%$ vs.\ $43.8\%$), indicating that even per-writer position guidance does not substantively expand the set of human sub-arguments reachable from LLM outputs.

\begin{table*}[t]
\centering\small
\renewcommand{\arraystretch}{1.20}
\setlength{\tabcolsep}{8pt}
\begin{tabular}{lcc}
\toprule
\textbf{Pair} & \textbf{Strict} & \textbf{Loose} \\
\midrule
Humans $\leftrightarrow$ \default       & $\phantom{0}9.7\%$ & $43.8\%$ \\
Humans $\leftrightarrow$ \persona       & $12.8\%$ & $42.9\%$ \\
\bottomrule
\end{tabular}
\caption{\textbf{Symmetric pool-to-pool recovery between humans and LLMs.} Macro-averaged $r$ over $16$ shared-main-argument cohorts, common-$m$ subsampled per cohort. Strict counts \texttt{equivalent} only; loose counts \texttt{equivalent} or \texttt{strong\_overlap}.}
\label{tab:app-cross-group-pool}
\end{table*}

\textbf{(ii) Per-family recovery from humans.} For each LLM family $f$, we compute the recovery of cluster-human sub-arguments into family $f$'s essay pool: $\text{Rec}_f = \text{avg}_i R(E_i \to P_f)$, where $E_i$ ranges over the cluster humans in a cohort and $P_f$ is the union of family $f$'s essays in that cohort under the given condition. \S\ref{tab:app-per-family-recovery} reports per-family recovery for \default (one essay per family per cluster-human seed per cohort) and \persona (one essay per family per cluster human writer per cohort). Recovery is modestly higher under \persona ($44$--$47\%$ loose) than \default ($37$--$44\%$ loose), and the spread across families is narrow in both conditions; no family is qualitatively closer to human reasoning than the others.

\begin{table*}[t]
\centering\small
\renewcommand{\arraystretch}{1.20}
\setlength{\tabcolsep}{8pt}
\begin{tabular}{lcccc}
\toprule
& \multicolumn{2}{c}{\textbf{\default}} & \multicolumn{2}{c}{\textbf{\persona}} \\
\cmidrule(lr){2-3}\cmidrule(lr){4-5}
\textbf{LLM family} & Strict & Loose & Strict & Loose \\
\midrule
GPT       & $\phantom{0}6.8\%$ & $43.6\%$ & $10.6\%$ & $44.1\%$ \\
Claude    & $\phantom{0}9.2\%$ & $40.0\%$ & $13.6\%$ & $44.8\%$ \\
Gemini    & $10.4\%$ & $37.3\%$ & $15.3\%$ & $46.5\%$ \\
Minimax   & $\phantom{0}9.7\%$ & $39.6\%$ & $13.2\%$ & $45.0\%$ \\
DeepSeek  & $\phantom{0}9.2\%$ & $40.8\%$ & $12.6\%$ & $45.8\%$ \\
\bottomrule
\end{tabular}
\caption{\textbf{Per-family recovery of cluster-human sub-arguments.} For each LLM family $f$, mean fraction of a cluster human's sub-arguments that have an \texttt{equivalent} (strict) or \texttt{equivalent}/\texttt{strong\_overlap} (loose) match in family $f$'s essay pool, averaged over cluster humans and then over the $16$ shared-main-argument cohorts.}
\label{tab:app-per-family-recovery}
\end{table*}

\paragraph{Cross-stance sub-argument reuse.}
\label{app:cross-stance-sampling}
As a stricter test of sub-argument collapse, we additionally analyze cross-stance sub-argument reuse on a subset of NYT-Room-for-Debate binary cohorts that show stance variation in both writer groups. Across $26$ such cohorts, \default LLMs reuse sub-arguments across opposite stances at $12.7\%$, roughly twice the human rate of $6.5\%$. Blinded coding shows LLM cross-stance overlap concentrates on functionalist assessments and definitional claims, whereas human overlap is more often grounded in causal-mechanism diagnoses.

\textbf{Cohort selection.} Starting from the $250$ binary cohorts for which essays have stance labels (\autoref{sec:content}), we require, for both the human pool and the \diversified LLM pool, at least two essays labeled \texttt{strong\_support} \emph{and} at least two essays labeled \texttt{strong\_oppose}. We further require that every one of the five LLM families contribute at least one strong-stance essay on each side, otherwise the family-balanced sampling step below is not possible. The intersection of these two criteria yields $26$ cohorts, spanning debates across politics, science, technology, sports, religion, and culture.

\textbf{Essay sampling.} For the LLM pool we sample one \diversified essay per family per stance per cohort (seed $42$), giving exactly $5$ \texttt{strong\_support} and $5$ \texttt{strong\_oppose} \diversified essays per cohort ($260$ \diversified essays total). This balances the LLM pool across families so the analysis does not depend on which family is most prolific in any given debate. For the human pool we keep every strong-stance essay (mean $4.85$/cohort, range $4$--$6$; $126$ humans total across the $26$ cohorts). Per-essay sub-argument counts are similarly tight in both groups (humans: mean $4.50$, median $4$, range $[3,6]$; \diversified: mean $4.50$, median $4$, range $[3,6]$).

\textbf{Pair construction.} Within each cohort we form all within-group essay pairs (humans--humans or \diversified--\diversified; cross-group pairs are skipped) and classify each by stance combination: \texttt{SS}, \texttt{OO}, or \texttt{SO}. Same-essay sub-argument pairs are excluded. Final essay-pair counts across the $26$ cohorts appear in \S\ref{tab:app-cross-stance-pairs}. For each of the $1{,}420$ inter-essay pairs we label every cross-essay sub-argument pair on the four-label scale (\autoref{sec:content}), yielding $27{,}439$ sub-argument pair judgments using the debate-question context prompt; model-call settings are reported in \S\ref{tab:model-hyperparameters}, and the sub-argument prompt is in \S\ref{app:prompts-content}.

\textbf{Symmetric reuse score $r(A,B)$.} For each within-group essay pair $(A,B)$ we compute the symmetric pair-level reuse rate
\[
r(A,B) = \tfrac{1}{2}\left(r_{A \to B} + r_{B \to A}\right),
\]
where $r_{A \to B}$ is the share of $A$'s sub-arguments that match at least one sub-argument in $B$, and $r_{B \to A}$ is defined symmetrically. This normalizes for differences in $|A|$ and $|B|$. A sub-argument is counted as ``reused'' if it matches some sub-argument in the other essay with relation \texttt{equivalent} or \texttt{strong\_overlap}. For each (writer-group, stance-combination) we take the per-cohort mean of $r$, then macro-average across the $26$ cohorts so each cohort contributes equally regardless of essay count. $95\%$ confidence intervals are computed by cluster bootstrap at the cohort level ($1{,}000$ resamples).

\textbf{Loose-threshold result.} The loose-threshold ($\{\texttt{equivalent},\,\texttt{strong\_overlap}\}$) macro-averaged rates are reported in \autoref{tab:app-stance-reuse}. Both groups drop sharply from same-stance to opposite-stance reuse; the human opposite-stance interval excludes any substantial cross-stance transfer, and the LLM cross-stance rate is roughly twice the human rate.

\begin{table*}[t]
\centering\small
\renewcommand{\arraystretch}{1.2}
\setlength{\tabcolsep}{6pt}
\begin{tabular*}{\textwidth}{@{\extracolsep{\fill}}lccc@{}}
\toprule
& \multicolumn{2}{c}{\textbf{same-stance}} & \textbf{opposite-stance} \\
\cmidrule(lr){2-3}
\textbf{group} & \texttt{SS} & \texttt{OO} & \texttt{SO} \\
\midrule
human  & $40.5_{\,\text{\scriptsize $[29.8,\,51.1]$}}$ & $35.0_{\,\text{\scriptsize $[26.2,\,44.3]$}}$ & $\phantom{0}6.5_{\,\text{\scriptsize $[4.0,\,9.0]$}}$ \\
LLM    & $66.3_{\,\text{\scriptsize $[59.1,\,72.9]$}}$ & $60.0_{\,\text{\scriptsize $[53.1,\,66.9]$}}$ & $12.7_{\,\text{\scriptsize $[9.9,\,15.8]$}}$ \\
\bottomrule
\end{tabular*}
\caption{\textbf{Cross-stance sub-argument reuse (loose threshold).} Macro-averaged symmetric pair-level reuse rate $r(A,B)$ (\%, counting \texttt{equivalent} or \texttt{strong\_overlap} matches) for within-group essay pairs across $26$ NYT binary cohorts. Subscripts give $95\%$ confidence intervals from cohort-level cluster bootstrap. Both groups drop sharply from same-stance to opposite-stance reuse; the human opposite-stance interval excludes any substantial cross-stance transfer.}
\label{tab:app-stance-reuse}
\end{table*}

\textbf{Strict-equivalence variant ($S = 1.0$).} The strict variant counts a sub-argument as reused only at exact equivalence (\texttt{equivalent}-only edges). The macro-averaged rates under this stricter threshold are reported in \S\ref{tab:app-cross-stance-strict}; both groups fall well below the loose-threshold rates, and both cross-stance rates fall below $1\%$, but the LLM rate remains above the human rate at every stance combination.

\textbf{Per-pair categorization.} We categorize cross-stance sub-argument pairs labeled \texttt{equivalent} or \texttt{strong\_overlap} in the $26$-cohort subset. Counts: $437$ LLM--LLM cross-stance pairs vs.\ $60$ human--human cross-stance pairs, a $7.3\times$ ratio in absolute counts that maps to the $\approx 2\times$ normalized rate in \autoref{tab:app-stance-reuse} after per-essay-size and per-cohort normalization.

\emph{Pass 1: per-pair characterization.} For each cross-stance pair, the judge sees the debate question, the support-side phrasing, and the oppose-side phrasing, and is asked to output (i)~a $2$--$5$ word topic-agnostic TYPE label characterizing the epistemic form of the shared content (e.g., empirical observation, structural diagnosis, normative principle, definitional claim, reform proposal), and (ii)~a one-sentence rationale for why this kind of statement can attach to either side of any debate. The prompt explicitly forbids topic-bound labels (e.g., ``jurisdictional principle'' for drug-enforcement debates) and supplies a candidate list of generic statement types.

\emph{Pass 2: emergent clustering.} The $437$ LLM-side TYPE labels (and separately, the $60$ human-side TYPE labels) are passed to a single follow-up call that clusters semantically equivalent labels into $3$--$6$ emergent categories, names each, defines it in one sentence, and lists members by index. The LLM-side clustering yields five categories (causal-mechanism diagnosis, functionalist assessment, normative principle, definitional / categorical boundary, structural reform proposal); the human-side clustering produces an analogous but coarser scheme. Cross-group category shares (LLM $33/19/18/17/13\%$; human $45/13/18/5/18\%$) are computed by manually aligning the two schemes. All judging uses \texttt{gemini-3-flash-preview} at temperature $0.2$, minimal reasoning effort. Representative cross-stance reuse pairs are shown in \autoref{tab:app-cross-stance-primitives}.

\begin{table*}[t]
\centering\small
\renewcommand{\arraystretch}{1.2}
\setlength{\tabcolsep}{6pt}
\begin{tabular*}{\textwidth}{@{\extracolsep{\fill}}lccc@{}}
\toprule
& \multicolumn{2}{c}{\textbf{same-stance}} & \textbf{opposite-stance} \\
\cmidrule(lr){2-3}
\textbf{group} & \texttt{SS} & \texttt{OO} & \texttt{SO} \\
\midrule
human  & $3.7_{\,\text{\scriptsize $[1.0,\,7.3]$}}$  & $4.3_{\,\text{\scriptsize $[1.0,\,8.5]$}}$  & $0.5_{\,\text{\scriptsize $[0.0,\,1.2]$}}$ \\
LLM    & $18.5_{\,\text{\scriptsize $[14.0,\,24.0]$}}$ & $16.0_{\,\text{\scriptsize $[12.5,\,19.5]$}}$ & $0.7_{\,\text{\scriptsize $[0.3,\,1.2]$}}$ \\
\bottomrule
\end{tabular*}
\caption{\textbf{Strict-equivalence variant ($S = 1.0$) of \autoref{tab:app-stance-reuse}.} Macro-averaged symmetric reuse $r$ (\%) when a sub-argument is counted as reused only at \texttt{equivalent} (not \texttt{strong\_overlap}). Subscripts are $95\%$ cluster-bootstrap CIs.}
\label{tab:app-cross-stance-strict}
\end{table*}

\begin{table*}[t]
\centering\small
\renewcommand{\arraystretch}{1.15}
\setlength{\tabcolsep}{8pt}
\begin{tabular*}{\textwidth}{@{\extracolsep{\fill}}lrrrr@{}}
\toprule
group & \texttt{SS} & \texttt{OO} & \texttt{SO} & total \\
\midrule
human-human   & $\phantom{0}52$  & $\phantom{0}47$  & $151$ & $\phantom{0,0}250$ \\
\diversified--\diversified & $260$ & $260$ & $650$ & $1{,}170$ \\
\midrule
total         & $312$ & $307$ & $801$ & $1{,}420$ \\
\bottomrule
\end{tabular*}
\caption{\textbf{Essay-pair counts for the cross-stance reuse analysis.} Within-group pairs across the $26$ binary cohorts after the two-stage selection and family-balanced \diversified sampling. Cross-group (human $\times$ \diversified) pairs are not used.}
\label{tab:app-cross-stance-pairs}
\end{table*}

\begin{table*}[t]
\centering\small
\renewcommand{\arraystretch}{1.3}
\setlength{\tabcolsep}{8pt}
\begin{tabular}{@{}p{0.47\linewidth}p{0.47\linewidth}@{}}
\toprule
\multicolumn{1}{c}{\textbf{Shared sub-argument}} & \multicolumn{1}{c}{\textbf{Essay-level position}} \\
\midrule

\multicolumn{2}{@{}l@{}}{\textbf{(A) Shared sub-claims}: both essays explicitly endorse the same supporting claim.} \\
\addlinespace[2pt]

\rowcolor{gray!12} \multicolumn{2}{@{}>{\centering\arraybackslash}p{\dimexpr 0.94\linewidth+2\tabcolsep\relax}@{}}{\textbf{Q. ``Breeding of a pedigreed dog creates genetic problems that a lovable mutt avoids?''}} \\
\multicolumn{2}{@{}p{\dimexpr 0.94\linewidth+2\tabcolsep\relax}@{}}{\textit{Shared sub-claim:} aesthetic breed standards drive the genetic health problems.} \\
\rowcolor{green!8} \textbf{Support:} ``Health testing by breeders is insufficient because \textbf{the breed standards themselves often reward dysfunctional anatomy} that causes lifelong suffering.''
& ``The practice of raising purebred dogs according to modern kennel club standards is ethically bankrupt because it prioritizes aesthetic traits over the health and physical functionality of the animals.'' \\
\rowcolor{red!8} \textbf{Oppose:} ``The genetic health issues associated with purebred dogs are \textbf{caused by aesthetic judging standards and poor breeder choices} rather than the inherent concept of a breed.''
& ``The practice of breeding pedigreed dogs should be reformed to prioritize health and function rather than abolished, as selective breeding is essential for maintaining the predictable traits required for specific working roles.'' \\
\midrule

\rowcolor{gray!12} \multicolumn{2}{@{}>{\centering\arraybackslash}p{\dimexpr 0.94\linewidth+2\tabcolsep\relax}@{}}{\textbf{Q. ``Western child labor standards should apply in developing countries?''}} \\
\multicolumn{2}{@{}p{\dimexpr 0.94\linewidth+2\tabcolsep\relax}@{}}{\textit{Shared sub-claim:} prohibition pushes children into more dangerous informal work.} \\
\rowcolor{green!8} \textbf{Support:} ``To be effective, labor prohibitions must be paired with corporate and governmental investment in education and financial support for families to prevent children from entering \textbf{more dangerous informal work}.''
& ``Western child labor standards should be applied globally because the right to a childhood is a universal principle that protects children from exploitation and long-term poverty regardless of a nation's economic status.'' \\
\rowcolor{red!8} \textbf{Oppose:} ``Blanket prohibitions on child labor can be counterproductive by \textbf{pushing children into more dangerous, unregulated informal work}.''
& ``Western child labor standards should not be unilaterally imposed on developing countries because they ignore historical context, economic realities, and the structural role Western nations play in creating global poverty.'' \\

\midrule\midrule

\multicolumn{2}{@{}l@{}}{\textbf{(B) Stance-neutral primitives}: both essays use the same observation or concept, applied differently to reach conclusions.} \\
\addlinespace[2pt]

\rowcolor{gray!12} \multicolumn{2}{@{}>{\centering\arraybackslash}p{\dimexpr 0.94\linewidth+2\tabcolsep\relax}@{}}{\textbf{Q. ``Doping should be allowed in sports?''}} \\
\multicolumn{2}{@{}p{\dimexpr 0.94\linewidth+2\tabcolsep\relax}@{}}{\textit{Shared primitive:} wealthy athletes or nations enjoy unequal pharmacological advantage.} \\
\rowcolor{green!8} \textbf{Support:} ``The current testing regime entrenches inequality by \textbf{favoring wealthy athletes and nations who can afford sophisticated masking agents and designer drugs}.''
& ``The current anti-doping regime should be replaced with a transparent, medically supervised system because the existing testing process is ineffective, unfair, and dangerous to athlete health.'' \\
\rowcolor{red!8} \textbf{Oppose:} ``Allowing doping would exacerbate global inequality by \textbf{favoring wealthy nations and athletes with access to superior pharmacological resources}.''
& ``Doping should remain prohibited in sports because legalization would lead to a dangerous pharmaceutical arms race that undermines the integrity of competition and endangers athletes of all ages.'' \\
\midrule

\rowcolor{gray!12} \multicolumn{2}{@{}>{\centering\arraybackslash}p{\dimexpr 0.94\linewidth+2\tabcolsep\relax}@{}}{\textbf{Q. ``The quest for energy efficiency pays off for the planet?''}} \\
\multicolumn{2}{@{}p{\dimexpr 0.94\linewidth+2\tabcolsep\relax}@{}}{\textit{Shared primitive:} efficiency gains are offset by high-consumption lifestyles, so real change requires living smaller.} \\
\rowcolor{green!8} \textbf{Support:} ``For affluent consumers, the benefits of efficient gadgets are often \textbf{negated by high-consumption lifestyles, necessitating a shift toward living in smaller spaces} and reducing overall consumption.''
& ``Energy efficiency should be treated as a matter of public policy and equity rather than a lifestyle choice for wealthy consumers, focusing on making low-carbon living affordable and accessible to everyone.'' \\
\rowcolor{red!8} \textbf{Oppose:} ``Meaningful climate action requires deliberate acts of restraint and lifestyle changes, such as \textbf{living in smaller spaces and driving less}, rather than substituting old products for new `green' ones.''
& ``The pursuit of energy-efficient consumer products is an ineffective climate strategy that distracts from the necessary goal of reducing overall consumption and addressing systemic drivers of emissions.'' \\

\bottomrule
\end{tabular}
\caption{\textbf{Representative cross-stance LLM sub-argument reuse pairs.} Section \textbf{(A)} shows cases where essays on opposite sides explicitly reuse the same supporting claim; section \textbf{(B)} shows cases where they reuse the same stance-neutral primitive but attach different value judgments or policy conclusions to it. In both cases, the shared sub-argument alone does not determine the essay's final position; stance emerges from how the broader main argument frames and integrates it.}
\label{tab:app-cross-stance-primitives}
\end{table*}

\subsection{Cluster ratio: multi-member $\rho$-distribution}
\label{app:cluster-ratio-distribution}

\begin{figure}[t]
\centering
\includegraphics[width=\linewidth]{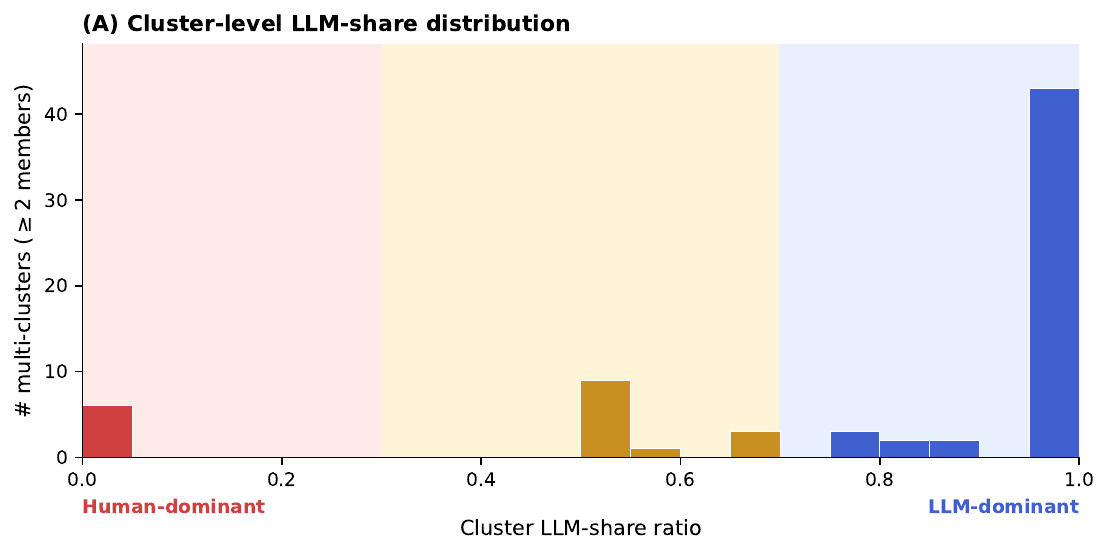}
\caption{\textbf{Distribution of multi-member cluster LLM-share $\rho$.} Histogram over the $69$ multi-member sub-argument clusters ($\geq 2$ members) in the $16$-cohort shared-main-argument subset. Cluster LLM-share $\rho = n_\mathrm{LLM} / (n_\mathrm{LLM} + n_\mathrm{Human})$ is already per-cluster normalized. Background bands mark the Human-dominant ($\rho \leq 0.3$), Mixed ($0.3 < \rho < 0.7$), and LLM-dominant ($\rho \geq 0.7$) regions used in \autoref{fig:cluster-ratio-overview}. The distribution is sharply asymmetric: $6$ Human-dominant, $13$ Mixed, $50$ LLM-dominant; no mixed cluster contains more humans than LLMs.}
\label{fig:app-cluster-ratio-distribution}
\end{figure}

\begin{figure}[t]
\centering
\includegraphics[width=\linewidth]{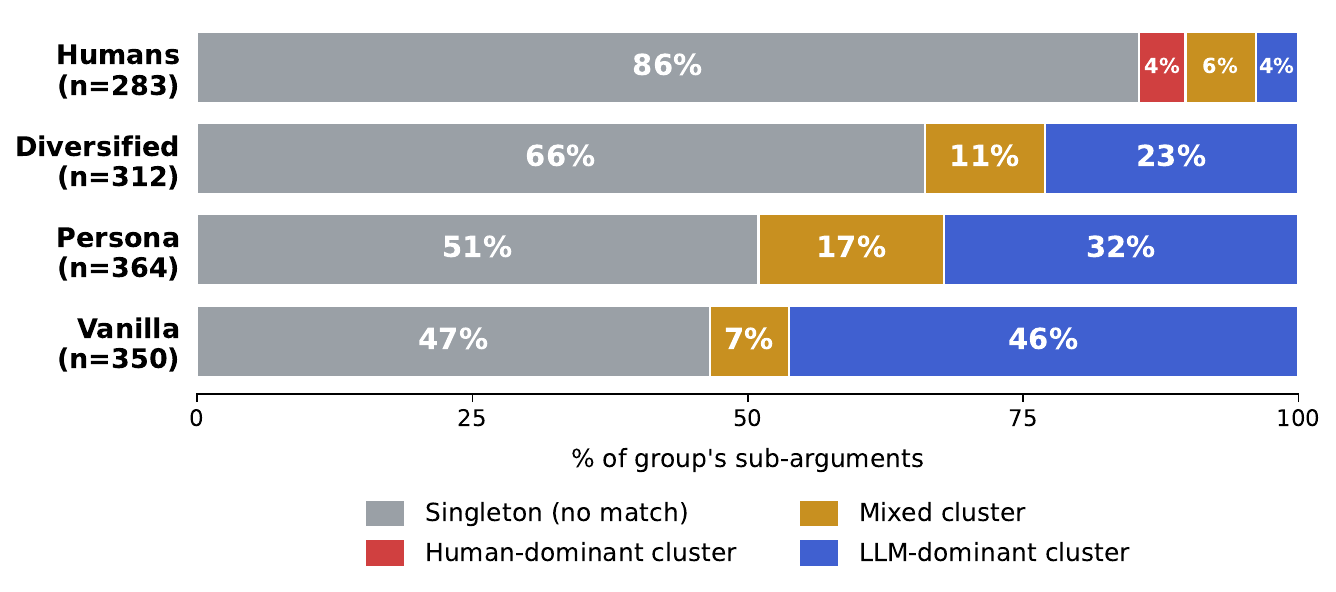}
\caption{\textbf{Per-condition sub-argument distribution across cluster regions.} Each LLM condition is clustered against the same human cluster separately; to prevent pool size from mechanically inflating clustering, each condition's pool is matched to \default (one essay per LLM family per cohort, mean over $10$ samplings). \diversified and \persona recover diversity only partially (singleton share $66\%$ and $51\%$), above \default ($47\%$) but well below humans ($86\%$).}
\label{fig:app-cluster-ratio-4cond}
\end{figure}

We characterize how each writer group's sub-arguments distribute across singleton and multi-member clusters. \autoref{fig:app-cluster-ratio-distribution} shows the histogram of multi-member cluster LLM-share $\rho$, and \autoref{tab:app-cluster-ratio-pergroup} reports the per-group breakdown by cluster region (singleton, human-dominant, mixed, LLM-dominant) that underlies \autoref{fig:cluster-ratio-overview}. \autoref{fig:app-cluster-ratio-4cond} extends this per-group breakdown to the \diversified and \persona conditions.

\begin{table*}[t]
\centering\small
\renewcommand{\arraystretch}{1.20}
\setlength{\tabcolsep}{5pt}
\begin{tabular}{lrrrrr}
\toprule
\textbf{Group} & \textbf{Singleton} & \textbf{H-dom} & \textbf{Mixed} & \textbf{LLM-dom} & \textbf{Total} \\
\midrule
Humans  & $242$ ($85.5\%$) & $12$ ($4.2\%$) & $18$ ($6.4\%$) & $11$ ($3.9\%$) & $283$ \\
\default & $163$ ($46.6\%$) & $0$ ($0.0\%$) & $25$ ($7.1\%$) & $162$ ($46.3\%$) & $350$ \\
\bottomrule
\end{tabular}
\caption{\textbf{Per-group distribution of sub-arguments across cluster regions.} For each group, count (and \% of group total) of sub-arguments that fall into: \emph{Singleton} (cluster size $= 1$), \emph{H-dom} (multi-member, $\rho \leq 0.3$), \emph{Mixed} ($0.3 < \rho < 0.7$), or \emph{LLM-dom} ($\rho \geq 0.7$). Underlies \autoref{fig:cluster-ratio-overview}. Humans concentrate in singletons; \default LLMs split roughly evenly between singletons and LLM-dominant clusters, with no contribution to human-dominant multi-clusters.}
\label{tab:app-cluster-ratio-pergroup}
\end{table*}

\subsection{Cluster ratio: qualitative analyses}
\label{app:cluster-ratio-qualitative}
We characterize the contents of the Human-dominant and LLM-dominant regions of the cluster-ratio distribution (\S\ref{sec:sub_arg}) through three complementary LLM-judge analyses. Together they answer three linked questions: what distinguishes human and LLM multi-member convergence, what remains uniquely human or uniquely LLM when no one else reuses it, and whether the same LLM pattern survives in the larger clusters that attract several independent essays. All three analyses operate within the same $16$-cohort shared-main-argument subset, use strict-threshold (\texttt{equivalent}-only) clusters, and present sub-arguments with the group identity (humans vs.\ LLMs) masked as ``Set A'' / ``Set B'' whenever a direct contrast is being made. The judge is \texttt{gemini-3-flash-preview} at temperature $0.4$ with $2500$ max output tokens.

\textbf{Cluster construction.} Within each cohort, we form sub-argument clusters from \texttt{equivalent} edges only (strict threshold) and characterize each multi-member cluster by its LLM share $\rho = n_\mathrm{LLM} / (n_\mathrm{LLM} + n_\mathrm{Human})$. We assign each cluster a region by $\rho$: \emph{Human-dominant} ($\rho \leq 0.3$), \emph{Mixed} ($0.3 < \rho < 0.7$), or \emph{LLM-dominant} ($\rho \geq 0.7$). For each multi-member cluster, the medoid is the member with the highest mean pairwise equivalence score to the other members.

\textbf{Analysis 1: blind within-cohort contrasts between multi-member convergence regions.} This analysis asks what human and LLM convergence look like when both sides contain a genuinely shared cluster. For every cohort with at least one multi-member ($\geq 2$) cluster in both Human-dominant and LLM-dominant regions ($6$ of $16$ cohorts qualify), we present all qualifying clusters from each region as a paired contrast (medoid only). Set A / Set B assignment is randomized per call (50/50). The judge is asked to identify $2$--$4$ recurring contrasts between the two sets, name each, and supply one representative phrase per side. One blinded contrast is produced per cohort.

\textbf{Analysis 2: blind contrasts between human-only and LLM-only singletons.} This analysis asks what each group contributes when no one else follows it. We therefore contrast the size-$1$ clusters (singletons): sub-arguments that belong only to a single essay and have no \texttt{equivalent} match anywhere in the cohort. Within each of the $16$ shared-main-argument cohorts ($242$ humans-only and $164$ LLMs-only singletons in total), we sample $\min(|S_H|, |S_L|, 20)$ singletons from each side, present them as Set A / Set B with randomized assignment, and request the same recurring-pattern / contrast output as Analysis 1. One blinded contrast is produced per cohort.

\textbf{Analysis 3: characterization of larger LLM-dominant clusters (size $\geq 3$).} The first two analyses cover the multi-member and singleton ends, but leave open whether the same picture survives in the larger LLM clusters that attract several distinct essays. We therefore extract every LLM-dominant cluster of size $\geq 3$ across the $16$ cohorts ($26$ clusters spanning $15$ cohorts), take each cluster's medoid, and ask the judge (a single call seeing all $26$ medoids together with their cohort tags) to identify recurring patterns and notable absences. This run is not blinded against a contrastive set; its goal is to characterize what survives convergence at higher cluster sizes, not to compare against humans.

\textbf{Synthesis.} The Pass-1 contrasts of Analyses 1 and 2 and the recurring-patterns output of Analysis 3 are the source of the qualitative claims in \S\ref{sec:sub_arg}: the same contrast emerges in all three analyses across debates from basketball rule design and cybersecurity to social networks, postpartum care, and coastal-property policy. Human arguments stay closer to concrete institutions, lived roles, and topic-specific constraints, whereas LLM arguments repeatedly move toward portable mechanism-level abstractions. \autoref{tab:human-llm-features} reports the five recurring sentence-level contrasts that survive this synthesis. Per-cohort Pass-1 outputs, the $26$ medoid characterizations, and an HTML viewer of all clusters by region are released with the code.

\section{Structure Appendix}
\label{app:structure}

\subsection{Paragraph-Level Taxonomies}
\label{app:structure-taxonomies}

\autoref{tab:argument} and \autoref{tab:discourse_mode} define the two paragraph-label layers used in the structure analysis. The argumentative-role layer captures what a paragraph does in the essay's argument, while the discourse-mode layer captures how the paragraph is written.

\begin{table*}[!t]
\centering
\small
\renewcommand{\arraystretch}{1.15}
\setlength{\tabcolsep}{4pt}
\begin{tabularx}{\textwidth}{@{}L{0.13\textwidth}YL{0.29\textwidth}L{0.20\textwidth}@{}}
\toprule
\textbf{Label} & \textbf{Definition} & \textbf{Example} & \textbf{Source} \\
\midrule
\texttt{None} & No role dominates; transitional, atmospheric, or pure setup. & A purely scenic anecdotal opener. & --- \\
\texttt{thesis} & Directly states the essay's central position. & ``The current immigration system is broken, and only comprehensive reform can repair it.'' & van Eemeren \& Grootendorst (2004) \\
\texttt{support} & Provides grounds, reasons, examples, or problem diagnosis that develops the case. & A paragraph laying out structural causes of a problem. & Walton (1996) \\
\texttt{reframing} & Recasts what the debate is really about; rejects misleading framing or substitutes a better lens. & ``The issue is not whether homeownership rates are down, but for whom declining ownership is harmful.'' & Entman (1993) \\
\texttt{counterclaim} & Voices an opposing position; the opposition is the dominant work. & ``Critics of this policy argue that it would lead to\ldots'' & Walton \& Krabbe (1995) \\
\texttt{rebuttal} & Defeats or undercuts the opposing position; the writer's response dominates. & A paragraph showing why the critics' worry doesn't survive scrutiny. & Toulmin (1958) \\
\texttt{concession} & Acknowledges that an opposing concern has real force or partial truth, without pivoting. & ``To be sure, this reform would impose real short-term costs on rural hospitals.'' & van Eemeren \& Grootendorst (2004) \\
\texttt{implication} & Draws out consequences from prior reasoning -- predictions, takeaways, ``therefore X follows.'' & ``This means second-order effects on adjacent industries will follow.'' & Pollock (1987) \\
\texttt{proposal} & Advocates a specific course of action -- what should be done, by whom, or under what conditions. & ``Cities should invest in public transit before subsidizing more parking.'' & Fairclough \& Fairclough (2012) \\
\bottomrule
\end{tabularx}
\caption{\textbf{Argument layer} (paragraph-level, multi-label, $|L|=8$ + \texttt{none}). Captures the paragraph's discourse-level argumentative role in the essay's progression.}
\label{tab:argument}
\end{table*}

\begin{table*}[!t]
\centering
\small
\renewcommand{\arraystretch}{1.15}
\setlength{\tabcolsep}{4pt}
\begin{tabularx}{\textwidth}{@{}L{0.16\textwidth}YL{0.30\textwidth}L{0.19\textwidth}@{}}
\toprule
\textbf{Label} & \textbf{Definition} & \textbf{Example} & \textbf{Source} \\
\midrule
\texttt{argumentation} & Claim-and-reason writing with explicit inferential force; reasoning drives sentence-to-sentence progression. & A paragraph that states a policy is flawed, then explains why the flaw matters. & Brooks \& Warren (1972) \\
\texttt{exposition} & Explanation, clarification, or factual information without dominant inferential force. & A paragraph explaining how a program or institution works. & Kinneavy (1971) \\
\texttt{narration} & Events or actions presented in temporal sequence; the organizing principle is time. & A paragraph recounting a sequence of events leading up to a decision. & Brooks \& Warren (1972) \\
\texttt{description} & Depiction of a scene, state, person, or place; organized spatially or by attributes. & A paragraph depicting conditions in a neighborhood or institution. & Kinneavy (1971) \\
\bottomrule
\end{tabularx}
\caption{\textbf{Discourse Mode layer} (paragraph-level, single-label, $|L|=4$). Captures how the paragraph is written, independent of its argumentative role.}
\label{tab:discourse_mode}
\end{table*}

\subsection{Full Structural Heatmaps}
\label{app:structure-heatmaps}

\begin{figure*}[t]
\centering
\includegraphics[width=\textwidth]{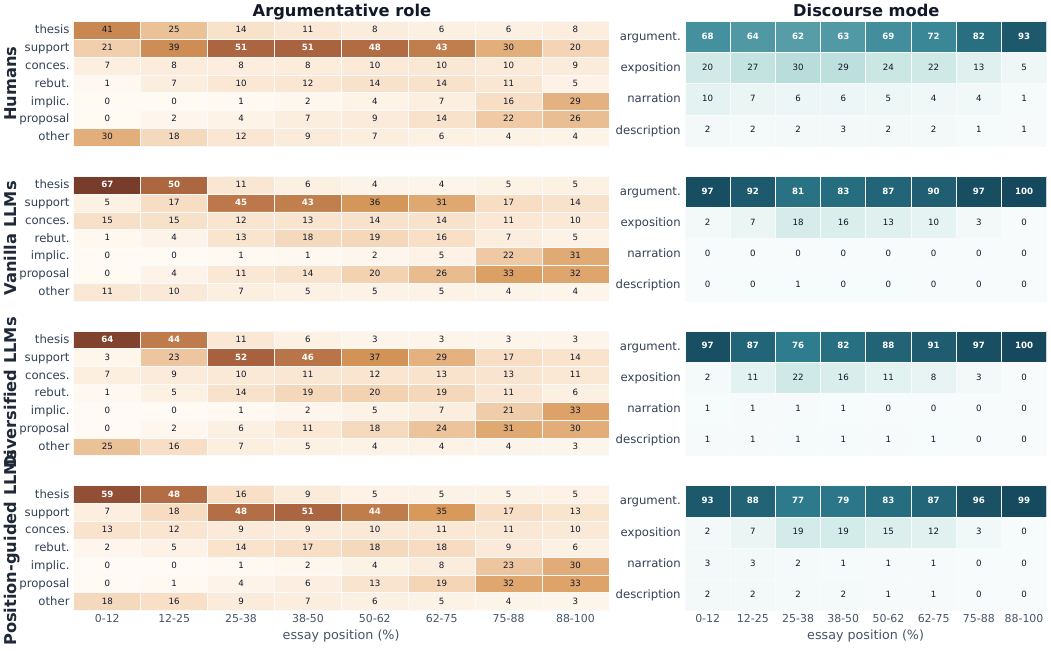}
\caption{\textbf{NYT structural heatmaps across all generation conditions.} Position-binned paragraph-label shares for human essays and LLM essays under \default, \diversified, and position-guided generation. The argument layer is multi-label, so cells report the share of role assignments within each position bin; the discourse layer is single-label, so cells report the share of paragraphs.}
\label{fig:structure-aggregate-heatmap}
\end{figure*}

\begin{figure*}[t]
\centering
\includegraphics[width=\textwidth]{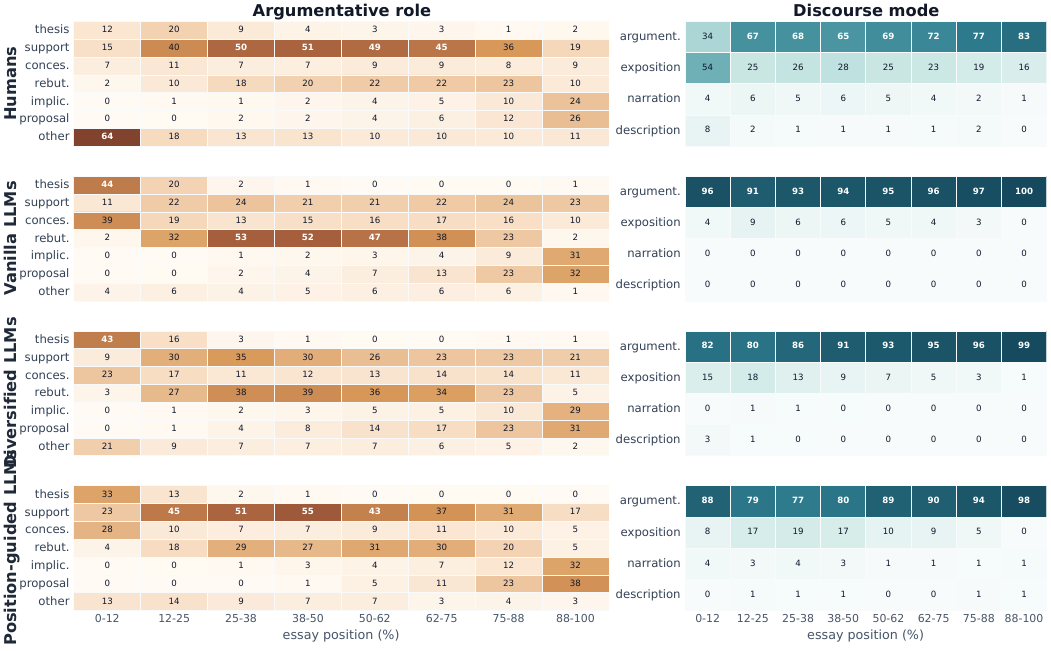}
\caption{\textbf{Boston Review structural heatmaps across all generation conditions.} Position-binned paragraph-label shares for human essays and LLM essays under \default, \diversified, and position-guided generation. The argument layer is multi-label, so cells report the share of role assignments within each position bin; the discourse layer is single-label, so cells report the share of paragraphs.}
\label{fig:structure-aggregate-heatmap-br}
\end{figure*}

\autoref{fig:structure-aggregate-heatmap} and \autoref{fig:structure-aggregate-heatmap-br} show the full position-binned label distributions for NYT and Boston Review. These figures expand the main-text structure discussion by showing all generation conditions for both label layers.

\subsection{Label-Flow Patterns}
\label{app:structure-flows}

\autoref{tab:structure-flow-patterns} and \autoref{tab:structure-flow-patterns-br} report selected paragraph-transition patterns for NYT and Boston Review. They support the main-text claim that human essays sustain supporting development more often, while LLM essays move from support toward proposal more quickly.

\begin{table*}[!htbp]
\centering
\small
\renewcommand{\arraystretch}{1.12}
\setlength{\tabcolsep}{5pt}
\begin{tabularx}{\textwidth}{@{}L{0.34\textwidth}rrY@{}}
\toprule
\textbf{Pattern} & \textbf{Human} & \textbf{Default LLM} & \textbf{Direction} \\
\midrule
\texttt{support} $\rightarrow$ \texttt{support} & 50.5 & 36.0 & Humans sustain supporting development more often. \\
\texttt{rebuttal} $\rightarrow$ \texttt{rebuttal} & 21.2 & 10.9 & Humans more often extend rebuttal across adjacent paragraphs. \\
\texttt{concession} $\rightarrow$ \texttt{support} & 47.1 & 40.6 & Humans more often return from concession to supporting development. \\
\texttt{support} $\rightarrow$ \texttt{proposal} & 12.3 & 29.4 & LLMs move from support to resolution more quickly. \\
\texttt{rebuttal} $\rightarrow$ \texttt{proposal} & 18.2 & 36.1 & LLMs more often close rebuttal with a proposal. \\
\texttt{concession} $\rightarrow$ \texttt{proposal} & 14.6 & 28.5 & LLMs more often convert concession into proposal. \\
\texttt{support} $\rightarrow$ \texttt{support} $\rightarrow$ \texttt{support} & 13.2 & 5.3 & Humans more often develop support over three paragraphs. \\
\texttt{thesis} $\rightarrow$ \texttt{support} $\rightarrow$ \texttt{proposal} & 1.6 & 5.8 & LLMs overuse a compact claim--support--proposal template. \\
\texttt{thesis} $\rightarrow$ \texttt{support} $\rightarrow$ \texttt{rebuttal} & 2.3 & 6.3 & LLMs overuse a compact claim--support--rebuttal template. \\
\bottomrule
\end{tabularx}
\caption{\textbf{Selected NYT paragraph-transition differences.} Bigram rows report $P(Y\text{ next}\mid X\text{ here})$ as percentages; trigram rows report occurrences per 100 paragraph triples. Default LLM values pool the five debate-only model conditions. Boston Review transition results are reported in \autoref{tab:structure-flow-patterns-br}.}
\label{tab:structure-flow-patterns}
\end{table*}

\begin{table*}[!htbp]
\centering
\small
\renewcommand{\arraystretch}{1.12}
\setlength{\tabcolsep}{5pt}
\begin{tabularx}{\textwidth}{@{}L{0.34\textwidth}rrY@{}}
\toprule
\textbf{Pattern} & \textbf{Human} & \textbf{Default LLM} & \textbf{Direction} \\
\midrule
\texttt{support} $\rightarrow$ \texttt{support} & 54.5 & 29.7 & Humans sustain supporting development more often. \\
\texttt{rebuttal} $\rightarrow$ \texttt{rebuttal} & 32.4 & 52.0 & LLMs extend rebuttal across adjacent paragraphs more often. \\
\texttt{concession} $\rightarrow$ \texttt{support} & 40.8 & 22.8 & Humans more often return from concession to supporting development. \\
\texttt{support} $\rightarrow$ \texttt{proposal} & 7.2 & 17.7 & LLMs move from support to resolution more quickly. \\
\texttt{rebuttal} $\rightarrow$ \texttt{proposal} & 10.5 & 13.8 & LLMs slightly more often close rebuttal with a proposal. \\
\texttt{concession} $\rightarrow$ \texttt{proposal} & 9.3 & 11.7 & LLMs slightly more often convert concession into proposal. \\
\texttt{support} $\rightarrow$ \texttt{support} $\rightarrow$ \texttt{support} & 14.7 & 3.0 & Humans more often develop support over three paragraphs. \\
\texttt{thesis} $\rightarrow$ \texttt{support} $\rightarrow$ \texttt{proposal} & 0.2 & 0.1 & This compact template is rare in Boston Review for both groups. \\
\texttt{thesis} $\rightarrow$ \texttt{support} $\rightarrow$ \texttt{rebuttal} & 0.9 & 2.7 & LLMs use this compact claim--support--rebuttal sequence more often. \\
\bottomrule
\end{tabularx}
\caption{\textbf{Selected Boston Review paragraph-transition differences.} Bigram rows report $P(Y\text{ next}\mid X\text{ here})$ as percentages; trigram rows report occurrences per 100 paragraph triples. Default LLM values pool the five debate-only model conditions.}
\label{tab:structure-flow-patterns-br}
\end{table*}

\section{Evaluator Analysis Details}
\label{app:evaluators}

\subsection{Evaluators}
\label{app:evaluator-details}

\autoref{tab:evaluator-descriptions} summarizes the six evaluators. Reward models (Skywork-Reward-V2-Llama-3.1-8B, ArmoRM-Llama3-8B-v0.1, LitBench-Creative-Writing-RM-3B) score the debate question as the user turn and the evaluated text as the assistant turn using each model's chat template (bf16; $4{,}096$-token truncation). Judge protocols run on \texttt{gemini-3.5-flash} at temperature $0$, and the 15 argument-quality ratings are averaged into a single score.

\begin{table*}[t]
\centering\small
\renewcommand{\arraystretch}{1.15}
\begin{tabularx}{\textwidth}{@{}llX@{}}
\toprule
\textbf{Evaluator} & \textbf{Type} & \textbf{What it measures} \\
\midrule
Skywork-Reward-V2 \citep{liu2025skywork} & Reward model & Scalar reward trained on human preference data; general response preference. \\
ArmoRM \citep{wang2024armorm} & Reward model & Preference score aggregated from reward dimensions labeled by humans and GPT-4. \\
MT-Bench \citep{zheng2023judging} & LLM judge & Pairwise verdict based on helpfulness, relevance, accuracy, depth, creativity, and detail. \\
WildBench \citep{lin2025wildbench} & LLM judge & Pairwise verdict against a checklist of debate-specific criteria, generated once per debate. \\
Argument quality \citep{wachsmuth-2017} & LLM judge & The 15 argument-quality dimensions from argumentation theory, each rated 1--3. \\
Writing preference \citep{fein2026litbench} & Reward model & Writing quality, trained on creative-writing preference data. \\
\bottomrule
\end{tabularx}
\caption{\textbf{Evaluators used in \S\ref{sec:evaluators}.}}
\label{tab:evaluator-descriptions}
\end{table*}
\subsection{Setup}
\label{app:evaluator-setup}

\paragraph{Data.}
Pools: $1{,}039$ human essays, $975$ \default{} (one per model$\times$debate), and $6{,}271$ \diversified{}. A main argument is \emph{shared} if another same-pool essay in the debate makes a same-mode argument, and \emph{unique} otherwise.

\paragraph{Same-argument pairs.}
We compare all $550$ human--\default{} pairs across $140$ debates whose main arguments are \texttt{equivalent}, using both full essays and extracted claims. Essays may occur in multiple pairs and debate-level aggregation absorbs this dependence. As a wording control, we also paraphrase each human claim sentence-by-sentence under a content-preservation instruction.

\paragraph{Shared-vs-unique comparisons.}
We compare shared- against unique-argument essays in the $102$ debates where both occur in both pools. A debate favors shared arguments when they win a majority of shared$\times$unique pairs. Pair-independent score evaluators use all pairs ($805$ human; $13{,}143$ \diversified{}; ties $=0.5$); pairwise judges use $K{=}3$ uniformly sampled pairs per debate (fixed seed), evaluated in both presentation orders (win $=1$, tie $=0.5$). We aggregate all contrasts by debate and apply two-sided binomial sign tests, dropping tied debates.

\subsection{Robustness and Controls}
\label{app:evaluator-robustness}

\paragraph{Claim-unit scoring.}
Scoring the extracted argument text alone preserves or strengthens the reward-model preference (Skywork $73.6$/$82.0\%$, ArmoRM $68.4$/$63.4\%$ of debates). The writing-preference model is unstable at this unit, and no claim-level variant exists for the pairwise judges or the rubric, whose protocols require full essays.

\paragraph{Paraphrase control.}
Human-claim paraphrasing leaves authorship preference largely intact: Skywork $84{\to}78\%$, ArmoRM $78{\to}85\%$, and MT-Bench $87{\to}78\%$ of debates (all $p<0.005$). The writing-preference model favors LLM claims only against paraphrased humans ($64\%$; original essays and claims n.s.), consistent with mild degradation from paraphrasing.

\clearpage
\onecolumn
\section{Prompts}
\label{app:prompts}

\subsection{Generation Prompts}
\label{app:prompts-generation}

\promptlisting{Default generation}{prompts/generation_default.md}
\promptlisting{Self-diversified generation}{prompts/generation_self_diversified.md}
\promptlisting{Position-guided generation}{prompts/generation_position_guided.md}

\subsection{Preprocessing Prompts}
\label{app:prompts-preprocessing}

\promptlisting{Topic tagging}{prompts/topic_tagging.md}
\promptlisting{Question-type tagging}{prompts/question_type_tagging.md}
\promptlisting{Sensitivity tagging}{prompts/sensitivity_tagging.md}
\promptlisting{Temporal-change tagging}{prompts/temporal_change_tagging.md}
\promptlisting{Persona and thesis extraction}{prompts/persona_extraction.md}

\subsection{Content Annotation Prompts}
\label{app:prompts-content}

\promptlisting{Toulmin-style argument extraction}{prompts/toulmin_extraction.md}
\promptlisting{Main-argument pairwise judge}{prompts/main_argument_pairwise_judge.md}
\promptlisting{Sub-argument pairwise judge}{prompts/sub_argument_pairwise_judge.md}
\promptlisting{Stance-axis extraction}{prompts/stance_axis_extraction.md}
\promptlisting{Stance labeling}{prompts/stance_labeling.md}
\promptlisting{Cluster-ratio contrast: multi-member regions (A1)}{prompts/cluster_contrast_a1.md}
\promptlisting{Cluster-ratio contrast: human-only vs LLM-only singletons (A2)}{prompts/cluster_contrast_a2.md}
\promptlisting{Cluster-ratio characterization: larger LLM-dominant clusters (A3)}{prompts/cluster_contrast_a3.md}

\subsection{Structure Annotation Prompts}
\label{app:prompts-structure}

\promptlisting{Argumentative-role annotation}{prompts/argument_role_annotation.md}
\promptlisting{Discourse-mode annotation}{prompts/discourse_mode_annotation.md}

\clearpage
\twocolumn

\end{document}